\documentclass[conference]{IEEEtran}
\IEEEoverridecommandlockouts
\usepackage{threeparttable} 
\usepackage{amsmath,amssymb,amsfonts}
\usepackage{algorithmic}
\usepackage{graphicx}
\usepackage{textcomp}

\usepackage[utf8]{inputenc}
\usepackage{xcolor}
\usepackage{listings}
\usepackage{caption}
\usepackage{subcaption}
\usepackage{threeparttable} 
\usepackage{siunitx}
\usepackage{multirow}
\usepackage{siunitx}
\usepackage{boldline} 
\usepackage{subcaption,mwe}
\usepackage{tabularx}
\usepackage{booktabs,chemformula}
\usepackage{mathtools}
\usepackage{makecell}
\usepackage{adjustbox}
\usepackage{url}
\usepackage{longtable}
\usepackage{mathrsfs}
\definecolor{lightblue}{RGB}{173,216,230}
\usepackage{booktabs}
\usepackage{siunitx}
\usepackage{makecell}
\usetikzlibrary{shapes.multipart}
\usepackage{tikz}
\usepackage{soul}
\usepackage{supertabular}
\usepackage{multicol}
\usepackage{threeparttable} 
\usetikzlibrary{calc} 
\usetikzlibrary{arrows, decorations.markings,positioning,backgrounds,shapes}
\definecolor{EMP}{HTML}{77DD77} 
\definecolor{NOR}{HTML}{06500C} 

\usepackage{threeparttable} 
\usepackage{booktabs} 
\usepackage{array}
\usepackage{multirow} 
\usepackage{caption} 
\usepackage{booktabs}
\usepackage{graphicx} 
\usepackage{booktabs} 
\usepackage{multirow} 
\usepackage{threeparttable} 
\usepackage{tabularx}
\usepackage{amsmath}
\usepackage{amssymb}
\usepackage{lipsum}
\usepackage{tikz}
\usepackage{soul}
\usetikzlibrary{calc} 
\usepackage[T1]{fontenc}
\usepackage[utf8]{inputenc}
\usepackage{graphicx}
\usepackage[british]{babel}
\usetikzlibrary{arrows, decorations.markings,positioning,backgrounds,shapes}
\usepackage{tcolorbox}
\usepackage{xcolor} 

\definecolor{pastelblue}{HTML}{aec6cf}
\definecolor{grayframe}{HTML}{d8d8d8}
\definecolor{pastelgreen}{HTML}{77dd77}
\definecolor{whiteframe}{HTML}{ffffff}
\definecolor{pastelblue}{HTML}{aec6cf}
\definecolor{grayframe}{HTML}{d8d8d8}
\definecolor{pastelgreen}{HTML}{77dd77}
\definecolor{whiteframe}{HTML}{ffffff}
\definecolor{pastelpink}{HTML}{ffb6c1}
\definecolor{creamframe}{HTML}{ffffcc}
\definecolor{lightlavender}{HTML}{e6e6fa}
\definecolor{pastelgray}{HTML}{cfcfc4}
\definecolor{pastelyellow}{HTML}{fdfd96}
\definecolor{lightgrayframe}{HTML}{d3d3d3}
\definecolor{lighterpastelblue}{HTML}{cde0e5}
\definecolor{lighterlavender}{HTML}{f3f3fd}  
\definecolor{verylightpastelblue}{HTML}{e0f0f5}  
\definecolor{verylightpastelblue}{HTML}{e0f0f5}  
\definecolor{lightgrayframe}{HTML}{f0f0f0}       
\definecolor{verylightpastelgreen}{HTML}{a3e4a7} 
\definecolor{verylightwhiteframe}{HTML}{ffffff}  
\definecolor{verylightpastelpink}{HTML}{ffdce0}  
\definecolor{verylightcreamframe}{HTML}{ffffe6}  
\definecolor{verylightlavender}{HTML}{fbfaff}    
\definecolor{verylightpastelgray}{HTML}{f0f0f0}  
\definecolor{verylightpastelyellow}{HTML}{ffffc2} 
\definecolor{verylightlightgrayframe}{HTML}{f5f5f5} 
\usepackage{colortbl}
\usepackage{listings}
\usepackage{xcolor}
\definecolor{burgundy}{RGB}{128,0,32}
\definecolor{lightburgundy}{RGB}{171,76,98}
\lstdefinestyle{SQLStyle}{
    language=SQL,
    basicstyle={\fontsize{5.5}{5.5}\ttfamily\selectfont},
    breaklines=true,
    breakatwhitespace=true,
    showstringspaces=false,
    keywordstyle=\color{blue},
    stringstyle=\color{lightburgundy},
    commentstyle=\color{green!60!black},
    numbers=none,
    tabsize=2,
    showtabs=false,
    frame=none,
    xleftmargin=0pt,
    aboveskip=5pt,
    belowskip=5pt
}
\lstdefinestyle{CypherStyle}{
    language=SQL, 
    basicstyle={\fontsize{5}{5}\ttfamily\selectfont},
    breaklines=true,
    breakatwhitespace=true,
    showstringspaces=false,
    keywordstyle=\color{blue},
    stringstyle=\color{black},
    commentstyle=\color{green!60!black},
    numbers=none,
    tabsize=2,
    showtabs=false,
    frame=none,
    xleftmargin=0pt,
    aboveskip=5pt,
    belowskip=5pt,
    morekeywords={MATCH, RETURN, CREATE, MERGE, WHERE, DELETE, WITH, OPTIONAL, DETACH, SET, LIMIT, ORDER, BY, SKIP, UNWIND, CALL}, 
    morestring=[b]", 
    moredelim=[s][\color{lightburgundy}]{-}{->}, 
    moredelim=[s][\color{lightburgundy}]{<}{-},  
    moredelim=[s][\color{lightburgundy}]{-[}{]-}, 
}
\lstdefinelanguage{json}{
  basicstyle=\fontsize{6}{6}\selectfont\ttfamily,
  showstringspaces=false,
  breaklines=true,    
  breakatwhitespace=true,
  literate=
    *{0}{{{\color{blue}0}}}{1}
     {1}{{{\color{blue}1}}}{1}
     {2}{{{\color{blue}2}}}{1}
     {3}{{{\color{blue}3}}}{1}
     {4}{{{\color{blue}4}}}{1}
     {5}{{{\color{blue}5}}}{1}
     {6}{{{\color{blue}6}}}{1}
     {7}{{{\color{blue}7}}}{1}
     {8}{{{\color{blue}8}}}{1}
     {9}{{{\color{blue}9}}}{1}
     {:}{{{\color{red}:}}}{1}
     {,}{{{\color{red},}}}{1}
     {\{}{{{\color{brown}\{}}}{1}
     {\}}{{{\color{brown}\}}}}{1}
     [{{{\color{brown}[}}}{1}
     ]{{{\color{brown}]}}}{1},
}

\usepackage{cite}
\usepackage{amsmath,amssymb,amsfonts}
\usepackage{algorithmic}
\usepackage{graphicx}
\usepackage{textcomp}
\usepackage{xcolor}
\usepackage{cite}
\usepackage{amsmath,amssymb,amsfonts}
\usepackage{algorithmic}
\usepackage{graphicx}
\usepackage{textcomp}
\usepackage{caption}
\usepackage{subcaption}
\usepackage{multirow}
\usepackage{boldline} 
\usepackage{subcaption,mwe}
\usepackage{soul}
\usepackage{tabularx}
\usepackage{booktabs,chemformula}
\usepackage{mathtools}
\usepackage{makecell}
\usepackage{adjustbox}
\usepackage{makecell}
\usepackage{threeparttable} 
\usepackage{booktabs} 
\usepackage{array}
\usepackage{multirow} 
\usepackage{caption} 
\usepackage{booktabs}
\usepackage{amssymb}
\usepackage{listings}
\usepackage{graphicx} 
\usepackage{booktabs} 
\usepackage{multirow} 
\usepackage{threeparttable} 
\usepackage{tabularx}
\usepackage[a4paper, total={184mm,239mm}]{geometry}
\def\BibTeX{{\rm B\kern-.05em{\sc i\kern-.025em b}\kern-.08em
    T\kern-.1667em\lower.7ex\hbox{E}\kern-.125emX}}
\begin{document}

\title{ChipXplore: Natural Language Exploration of Hardware Designs and Libraries
\thanks{This work is partially supported by NSF grants 2350180 and 2453413.}}

\author{
Manar Abdelatty, Jacob K. Rosenstein, and Sherief Reda \\ Brown University, Providence, RI, USA}

\maketitle

\begin{abstract}

Hardware design workflows rely on Process Design Kits (PDKs) from different fabrication nodes, each containing standard cell libraries optimized for speed, power, or density. Engineers typically navigate between the design and target PDK to make informed decisions, such as selecting gates for area optimization or enhancing the speed of the critical path. However, this process is often manual, time-consuming, and prone to errors. To address this, we present ChipXplore, a multi-agent collaborative framework powered by large language models that enables engineers to query hardware designs and PDKs using natural language. By exploiting the structured nature of PDK and hardware design data, ChipXplore retrieves relevant information through text-to-SQL and text-to-Cypher customized workflows. The framework achieves an execution accuracy of 97.39\% in complex natural language queries and improves productivity by making retrieval 5.63× faster while reducing errors by 5.25× in user studies. Compared to generic workflows, ChipXplore's customized workflow is capable of orchestrating reasoning and planning over multiple databases, improving accuracy by 29.78\%. ChipXplore lays the foundation for building autonomous agents capable of tackling diverse physical design tasks that require PDK and hardware design awareness.

\end{abstract}


\section{Introduction}

At the core of semiconductor workflows lies a critical component, the Process Design Kit (PDK). PDKs provide comprehensive standard cell libraries for synthesizing abstract hardware designs into manufacturable chips. These libraries are optimized for specific metrics such as speed, density, or power, with detailed files on cell timing across process corners, physical layout data, and metal stack properties. Traditionally, hardware engineers manually navigate this complex landscape, parsing extensive library files with thousands of cells and attributes to locate relevant information. This process is time-consuming and error-prone, highlighting the need for automated tools that efficiently manage and utilize PDK and hardware design information.

Large Language Models (LLMs) have emerged as powerful assistants in hardware design, aiding tasks such as Verilog code generation~\cite{Takur2022vgen,pearce2020dave,blocklove2023chip,lu2024rtllm,liu2023rtlcoder,Liu2023verilog,liu2023chipnemo}, optimization~\cite{yao2024rtlrewriter}, RTL bug fixing~\cite{tsai2023rtlfixer}, EDA tool scripting~\cite{wu2024chateda}, and documentation question-answering~\cite{pu2024customized}. However, their application for querying PDK and design information remains underexplored. LLMs could enhance \emph{engineer-PDK-design} interactions through natural language interfaces and automated data retrieval. However, LLMs lack inherent knowledge of PDK data. Although domain adaptive pre-training could address this~\cite{gururangan2020don}, it is impractical due to extensive and costly training on diverse PDK datasets and the need for frequent retraining to stay up-to-date with new PDK versions.
\begin{figure}[!t]
    \centering
    \includegraphics[width=\columnwidth,trim={0.2cm 2.6cm 0.3cm 0.8cm}]{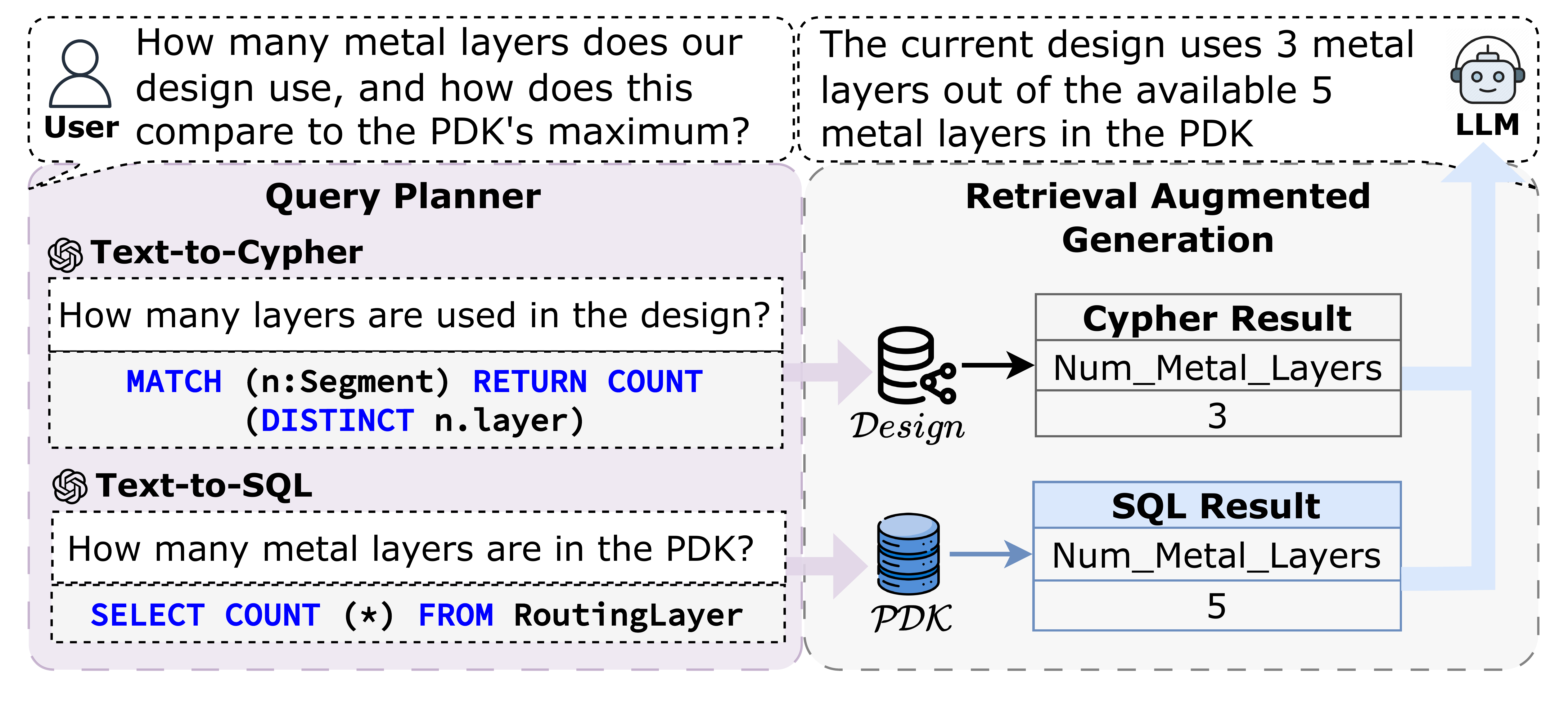}
    \vspace{-5mm}
    \caption{Overview of \textbf{\emph{ChipXplore}}. The framework converts a user question to $SQL$ and $Cypher$ to retrieve PDK and design data, then generates natural language answers using an LLM.}
    \vspace{-3.mm}
    \label{fig:plot1}
\end{figure}

Retrieval Augmented Generation (RAG) provides a flexible and maintainable approach to introduce PDK and design awareness into LLM by integrating responses in up-to-date external knowledge sources without model retraining~\cite{lewis2020retrieval,chen2024benchmarking,Wenqi2024asurvey}. RAG supports querying unstructured (e.g., text)\cite{siriwardhana2023improving}, semi-structured (e.g., JSON), and structured data (e.g., relational and graph databases)\cite{pourreza2024din,wang2023mac,li2024can,zhong2020semantic,pourreza2024din,gao2023text,dong2023c3,feng2023robust,Tran2024robust}. For unstructured data, RAG uses semantic similarity to retrieve relevant information~\cite{Pankaj2024robust}; for semi-structured data, it combines semantic and structural information. For structured data, RAG dynamically generates queries based on user input, such as text-to-SQL for relational databases~\cite{wang2023mac,li2024can,zhong2020semantic,pourreza2024din,gao2023text,dong2023c3} and text-to-Cypher for graph databases~\cite{Tran2024robust}.

Process Design Kits (PDKs), with their structured format, integrate seamlessly into relational databases, while hardware designs in Design Exchange Format (DEF)~\cite{def_lef_manual} are well suited to graph databases. Converting standard EDA file formats into databases offers multiple advantages. First, it allows LLMs to precisely retrieve data by dynamically generating database queries based on user input. Second, it enables the secure handling of proprietary PDK and design data: proprietary LLMs can be employed solely to generate database queries, while trusted local LLMs interpret the raw query results.

In light of this, we propose \emph{ChipXplore}\footnote{https://github.com/scale-lab/ChipXplore}, shown in Fig.~\ref{fig:plot1}, an LLM-powered framework to interact with hardware design information and PDKs using natural language. \emph{ChipXplore} harnesses the power of RAG, text-to-SQL, and text-to-Cypher conversions to provide an accurate interface for accessing PDK and design information.  To the best of our knowledge, \emph{ChipXplore} is the first framework that addresses the task of LLM-assisted navigation of PDKs and hardware designs.

Our contributions are summarized as follows:
\begin{itemize}
    \item We propose \emph{ChipXplore}, a customized multi-agent LLM-based workflow that streamlines natural language interactions with hardware designs and Process Design Kits (PDKs), enhancing hardware design engineers' efficiency. 

    \item We design relational and graph database schemas tailored for storing PDK and hardware design data, respectively, facilitating precise LLM-based retrieval through the dynamic generation of SQL and Cypher queries based on natural language input.  
    
    \item Our experimental results show that \emph{ChipXplore} effectively handles a wide range of complex user queries, achieving an overall execution accuracy of 97.39\%.  

    \item Compared to generic workflows~\cite{wang2023mac}, \emph{ChipXplore} customized workflow integrates reasoning across multiple heterogeneous databases, delivering a 29.78\% improvement in accuracy and underscoring the importance of specialized workflows for hardware design data management.  
    
    \item User studies reveal that \emph{ChipXplore} enhances productivity by making retrieval 5.63× faster and reducing human errors by 5.25×, highlighting its effectiveness in accelerating time-intensive tasks and improving design reliability.

\end{itemize}

This paper is organized as follows. Section~\ref{related_work} discusses related work. Section~\ref{framework} provides an overview of the \emph{ChipXplore} framework. Section~\ref{exp-results} presents experimental results. Finally, Section~\ref{conclusion} concludes the paper.

\section{Related Work}
\label{related_work}

\begin{figure*}[!h]
    \centering
    \includegraphics[width=\textwidth,trim={5cm 0.2cm 29cm 3cm}]{overview/overview_secure2.pdf}
    \vspace{-5mm}
\caption{Overview of the \textbf{\emph{ChipXplore}} multi-agent workflow, consisting of six LLM agents: (1) \emph{Planner} plans a sequence of actions based on user input; (2) \emph{Dispatcher} routes questions to relevant data sources in the $\mathcal{PDK}$ and $Design$ databases; (3) \emph{Selector} identifies relevant tables for $\mathcal{PDK}$ queries and nodes for $Design$ queries; (4) \emph{Query Generator} decomposes the user question into sub-questions to create $SQL$ and $Cypher$ queries; (5) \emph{Refiner} executes queries and fixes any syntax or logical errors; and (6) $Interpreter$ formulates final answer based on the raw database results.}

\label{fig:overview}
\vspace{-6mm}
\end{figure*}

The use of large language models (LLMs) for managing large-scale structured data has gained attention for their ability to expedite data retrieval and generate insights rapidly.  For relational databases, several frameworks have been proposed to optimize Retrieval Augmented Generation (RAG)-based SQL pipelines through in-context learning strategies. DIN-SQL\cite{pourreza2024din} introduced a chain-of-thought query decomposition strategy that decomposes SQL generation into smaller sub-problems. C3-SQL~\cite{dong2023c3} improved this with a zero-shot prompting technique, reducing token count compared to few-shot query decomposition. MAC-SQL\cite{wang2023mac} further advanced this by introducing a multi-agent framework with stages for table selection, query decomposition, and query refinement. Other methods investigated the use of fine-tuning to improve the performance of smaller models such as DTS-SQL~\cite{pourreza2024dts}, Chase-SQL~\cite{pourreza2024chase}, and Xiyan-SQL~\cite{gao2024xiyan} proposed a multiple finetuned query generators and selectors pipeline for increasing the likelihood of generating a correct query.

In parallel, LLMs have also been explored for querying graph-structured data. This approach, called GraphRAG~\cite{peng2024graph}, integrates graph databases into response generation by dynamically creating graph queries, such as Cypher for Neo4j ~\cite{neo4j_graph_database}, to retrieve relevant information. This method is used for querying data organized in a graph structure such as logical mind maps ~\cite{wu2025agentic} and knowledge graphs extracted from text ~\cite{xu2024retrieval, peng2024graph, wu2024medical, chen2024plan}.

Although these techniques have demonstrated the effectiveness of LLMs in querying structured data, they primarily focus on general-purpose relational and graph databases~\cite{yu2018spider,li2024can,lei2024spider}, often treating them in isolation. Their application to hardware design databases remains largely unexplored.  These methods are not optimized for the structured redundancy inherent in PDK and design data, where entities such as standard cells appear across multiple libraries, threshold voltages, and operating conditions. Additionally, existing approaches lack the ability to orchestrate reasoning across heterogeneous databases that require integrating both SQL (for PDK information) and Cypher (for design information). As a result, they cannot handle cross-database queries that require  reasoning over both PDK and hardware design data simultaneously. In addition, these approaches are not designed with security and scalability in mind, making them unsuitable for handling large scale proprietary data.  All these complexities underscore the pressing need for a customized workflow for hardware design databases. 

In this work, we address this gap by introducing \emph{ChipXplore}, a novel framework customized for hardware design databases. It enables cross-database reasoning and planning, handling complex queries that require integrating both PDK and hardware design data. By incorporating structured redundancy awareness into SQL and Cypher generation, \emph{ChipXplore} enables precise retrieval of complex hardware design data, accelerating design workflows and decision-making processes.

\section{ChipXplore Framework}
\label{framework}

Fig.~\ref{fig:overview} presents the \emph{ChipXplore} multi-agent workflow, which employs six  LLM agents, each designed for a specific function in the query processing pipeline. This section provides an overview of the proposed database schemas for storing PDK and design information, followed by a detailed explanation of each agent's functionality.

\subsection{LLM-Compatible Database Schema}\label{AA}

\begin{figure}[!t]
\centering
\begin{tikzpicture}[
    yscale=0.25, 
    relation/.style={
        rectangle split, 
        rectangle split parts=#1, 
        rectangle split part align=base, 
        draw, 
        anchor=center, 
        align=center, 
        text height=0.45mm, 
        text centered, 
        font=\fontsize{7}{5}\selectfont,
        rectangle split part fill={blue!7},
        inner sep=2pt
    },
    title/.style={
        font=\fontsize{7}{7}\selectfont
    }
]

\node (countrytitle) [title] {\textbf{OperatingConditions}};

\node [relation=7, rectangle split horizontal, anchor=north west, below=0.36cm of countrytitle.west, anchor=west] (opcond)
{\underline{Condition\_ID}
\nodepart{two} Name
\nodepart{three} Voltage
\nodepart{four} Temperature
\nodepart{five} Process
\nodepart{six} ...
\nodepart{seven} Standard\_Cell\_Library
};;1

\node [title, below=0.4cm of opcond.west, anchor=west] (cells) {\textbf{Cells}};

\node [relation=8, rectangle split horizontal, below=0.35cm of cells.west, anchor=west] (cells)
{\underline{Cell\_ID}%
\nodepart{two}  Name
\nodepart{three} Area
\nodepart{four} Leakage\_Power
\nodepart{five} Footprint
\nodepart{six} Waveform\_Fall
\nodepart{seven} ...
\nodepart{eight} Condition\_ID
};

\node [title, below=0.45cm of cells.west, anchor=west] (awardtitle) {\textbf{InputPins}};

\node [relation=7, rectangle split horizontal, below=0.35cm of awardtitle.west, anchor=west] (inputpins)
{\underline{Input\_Pin\_ID}%
\nodepart{two} Cell\_ID
\nodepart{three} Name
\nodepart{four} Clock
\nodepart{five} Capacitance
\nodepart{six} ...
\nodepart{seven} Rise\_Capacitance};

\node [title, below=0.45cm of inputpins.west, anchor=west] (outputpinstitle) {\textbf{OutputPins}};

\node [relation=7, rectangle split horizontal, below=0.35cm of outputpinstitle.west, anchor=west] (outputpins)
{\underline{Output\_Pin\_ID}%
\nodepart{two} Cell\_ID
\nodepart{three} Name
\nodepart{four} Function
\nodepart{five} Max\_Transition
\nodepart{six} ...
\nodepart{seven} Max\_Capacitance};

\node [title, below=0.45cm of outputpins.west, anchor=west] (timingtitletitle) {\textbf{TimingValues}};
\node [relation=5, rectangle split horizontal, below=0.35cm of timingtitletitle.west, anchor=west] (timingtables)
{\underline{Timing\_Value\_\_ID}%
\nodepart{two} Output\_Pin\_ID
\nodepart{three} Related\_Input\_Pin
\nodepart{four} ...
\nodepart{five} Fall\_Delay};

\draw  (cells.seven north) -- ++(0.0,0.6) -|  ++(-0cm,0.3) -| (opcond.one south) +(1.6, 0.9);

\draw  (inputpins.two north) -- ++(0,0.6)  -|  ++(-0cm,0.6) -| (cells.one south) +(0.8, 1.2);

\draw (outputpins.two north) -- ++(0,1.23) -|++(-2.2cm,4.7) -| (cells.one west) ++(0, 1.4);

\draw  (timingtables.two north) -- ++(0,1.2) -| (outputpins.one south) +(0.6, 0.8);


\end{tikzpicture}
\vspace{-1.1mm}
\caption{Liberty schema for timing data, showing relationships between operating conditions, cells, pins, and timing tables.}
\vspace{-2.4mm}
\label{fig:lib_schema}
\end{figure}

To enable LLM-based retrieval, we first convert the PDK files into a relational database, focusing on three key views: TechnologyLEF, Liberty, and LEF. The PDK schemas are designed to store information across multiple standard cell libraries and operating corners within a relational database. The Technology LEF schema captures details for routing layers (e.g., width, spacing rules, and resistance), via layers, and associated antenna ratios. The LEF schema stores the abstract physical representation of cells, such as cell dimensions, pin shapes, and obstruction layers, along with antenna effects. The Liberty schema, shown in Fig.~\ref{fig:lib_schema}, records the timing and electrical characteristics of cells, including tables for operating conditions, cell attributes, pin properties, and timing data.

Additionally, we define a schema to store hardware designs in DEF format~\cite{def_lef_manual} within a graph database. The design schema, shown in Fig.\ref{fig:hw_design_schema}, represents hardware design components as graph nodes, including pins, cells, nets, and segments, with semantic edges capturing the connectivity between these nodes. Together, these schemas enable efficient storage and retrieval of PDK and design data, allowing LLMs to retrieve information from both databases by dynamically generating SQL and Cypher queries based on user input.



\subsection{LLM Agents}\label{SCM} \emph{ChipXplore} comprises six LLM agents: \emph{Planner}, \emph{Dispatcher}, \emph{Selector}, \emph{Query-Generator}, \emph{Refiner}, and \emph{Interpreter}, which collaborate to ensure accurate and efficient extraction of data, facilitating PDK and design queries.

\textbf{\emph{Planner}}: The \emph{Planner} is a ReAct-style agent~\cite{yao2022react} responsible for orchestrating actions based on the user’s question. It determines how to interact with databases and other agents, structuring the process step by step. At each iteration \(i\), the planner selects an action \(A_{i}\) from three options: \textit{query\_design}, \textit{query\_pdk} to fetch information from a database or \textit{interpret} action to formulate the final answer based on the database results \(R_{i}\). \textit{interpret} actions can be delegated to a local trusted LLM to ensure secure handling of private PDK and design data. For the \textit{query} actions, the planner defines the question \(Q_i\) to be asked for the subsequent agents. Eq.~\ref{eq:planner_iterative} describes the Planner’s process, where \(f_{\text{\emph{planner}}}\) represents prompting the LLM agent \(\mathcal{M}\) with the user question \(Q\) and actions from prior steps \(\mathcal{A}_{i-1}\) to determine the next action.

\begin{figure}[!t]
\centering
\vspace{-2.3mm}
\begin{tikzpicture}
\definecolor{myorange}{RGB}{255,165,0}
\definecolor{myolivegreen}{RGB}{85,107,47}
\definecolor{myroyalblue}{RGB}{65,105,225}
\definecolor{myred}{RGB}{255,0,0}
\definecolor{mygoldenrod}{RGB}{218,165,32}
\definecolor{mypurple}{RGB}{128,0,128}
\definecolor{myteal}{RGB}{0,128,128}
\tikzset{
    pics/mynode/.style n args={4}{code={
        \node (#1) at (0,0) [draw=blue,text=black,font=\fontsize{7}{6}\selectfont, align=left,thick,minimum width=0.8cm,minimum height=1cm, rounded corners,label={[font=\fontsize{7}{7}\selectfont\bfseries,yshift=-2pt]#3}]{#4};
    }},
    myarrow/.style={->,>=latex,line width=0.7pt},
    pics/myh/.style n args={5}{code={
  \draw[myarrow, myorange] (#1) to[bend right=#3] node[midway,font=\fontsize{6}{5}\selectfont, rotate=#4,#5] {HAS\_PORT} (#2);
      }},
   pics/mycontains/.style n args={5}{code={
  \draw[myarrow, myolivegreen] (#1) to[bend left=#3] node[midway,font=\fontsize{6}{5}\selectfont, rotate=#4,#5] {CONTAINS\_CELL} (#2);
      }},
   pics/myrouted/.style n args={5}{code={
  \draw[myarrow, mypurple] (#1) to[bend left=#3] node[midway,font=\fontsize{6}{5}\selectfont, rotate=#4,#5] {ROUTED\_ON} (#2);
      }},
   pics/mycontainsnet/.style n args={5}{code={
  \draw[myarrow, myteal] (#1) to[bend right=#3] node[midway,font=\fontsize{6}{5}\selectfont, rotate=#4,#5] {CONTAINS\_NET} (#2);
      }},
   pics/myhaspin/.style n args={5}{code={
  \draw[myarrow, myroyalblue] (#1) to[bend left=#3] node[midway,font=\fontsize{6}{5}\selectfont, rotate=#4,#5] {HAS\_PIN} (#2);
      }},
   pics/mydrivesnet/.style n args={5}{code={
  \draw[myarrow, myred] (#1) to[bend left=#3] node[midway,font=\fontsize{6}{5}\selectfont, rotate=#4,#5] {DRIVES\_NET} (#2);
      }},
   pics/myconnectsto/.style n args={5}{code={
  \draw[myarrow, mygoldenrod] (#1) to[bend right=#3] node[midway,font=\fontsize{6}{5}\selectfont, rotate=#4,#5] {CONNECTS\_TO} (#2);
      }},
 }
    \matrix[row sep=-0.05cm, column sep={1.8cm,0cm,-0.1cm}] {
    \pic {mynode={design}{}{Design}{Name\\ Die\_Area\\ Stage \\ Xmin \\ .. \\ Ymax}}; &
    \pic {mynode={cell}{}{Cell}{Name\\ Instance\_Name\\  Orientation\\  Dynamic\_Power\\  Is\_In\_CLK \\  Is\_Sequential\\ ...\\ X0,Y0,X1,Y1 }}; &
    \pic {mynode={cellpin}{}{CellInternalPin}{Name\\ Direction\\ Transition\\ Slack\\ Rise\_arr\\ ...\\ Capacitance}}; & \\
    \pic {mynode={designport}{}{Port}{Name\\ Direction\\ Signal\_Type\\ Width\\ ...\\ Layer}}; &
    \pic {mynode={net}{}{Net}{Name\\ Signal\_Type\\ Capacitance\\ Resistance\\ Fanout\\ ... \\ Length}}; &
    \pic {mynode={segment}{}{Segment}{Layer\\ X0 \\ ... \\ Y1}}; & \\
 };
\pic {myh={design.south}{designport.north west}{89}{70}{above}};
\pic {mycontains={design.east}{cell.west}{0}{0}{above}};
\pic {myhaspin={cell.east}{cellpin.west}{0}{0}{above}};
\pic {mydrivesnet={cellpin.south}{net.north east}{-15}{10}{above}};
\pic {myconnectsto={net.north east}{cellpin.south}{15}{10}{below}};
\pic {myrouted={net.east}{segment.west}{0}{0}{above}};
\pic {mycontainsnet={design.south east}{net.north west}{0}{-14}{below}};
\pic {mydrivesnet={designport.east}{net.west}{20}{0}{above}};
\pic {myconnectsto={net.west}{designport.east}{-20}{0}{below}};
\end{tikzpicture}
\vspace{-6mm}
\caption{DEF schema for storing hardware design components in a graph database, illustrating semantic relationships between design, port, cell, pin, net, and segment nodes.}
\label{fig:hw_design_schema}
\vspace{-4mm}
\end{figure}

\vspace{-3mm}
\begin{equation}
\footnotesize
\langle (\mathcal{A}_i, \mathcal{I}_i) \rangle = f_{\text{\emph{planner}}}(\mathcal{Q}, \mathcal{A}_{i-1} \mid \mathcal{M})
\label{eq:planner_iterative}
\end{equation}


\begin{figure}[!t]
    \centering
    \begin{subfigure}{\linewidth}
        \centering
       \includegraphics[width=1\linewidth, trim={1.1cm 1.0cm 1.8cm 1cm}]{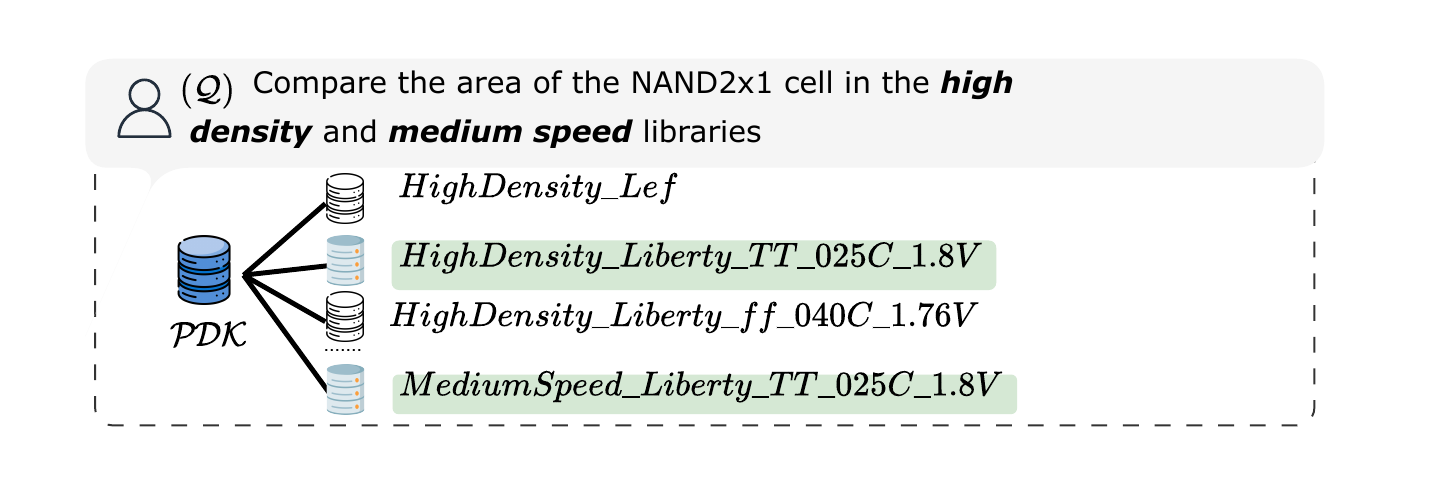}
        \vspace{-5mm}
        \caption{}
        \label{fig:study_a}
    \end{subfigure}
    \hfill
    \begin{subfigure}{0.49\linewidth}
        \centering
        \includegraphics[width=\linewidth]{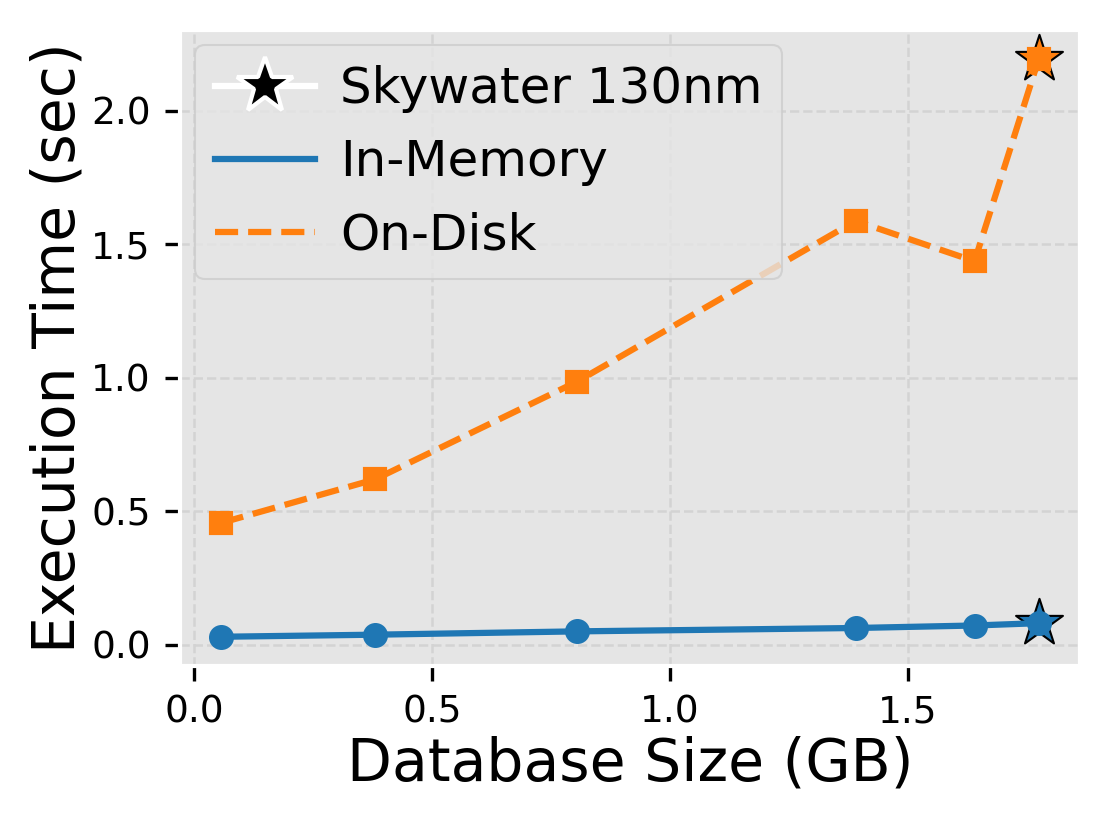}
        \vspace{-6mm}
        \caption{Sky130nm PDK}
        \label{fig:sky130}
    \end{subfigure}
    \begin{subfigure}{0.49\linewidth}
        \centering
        \includegraphics[width=\linewidth]{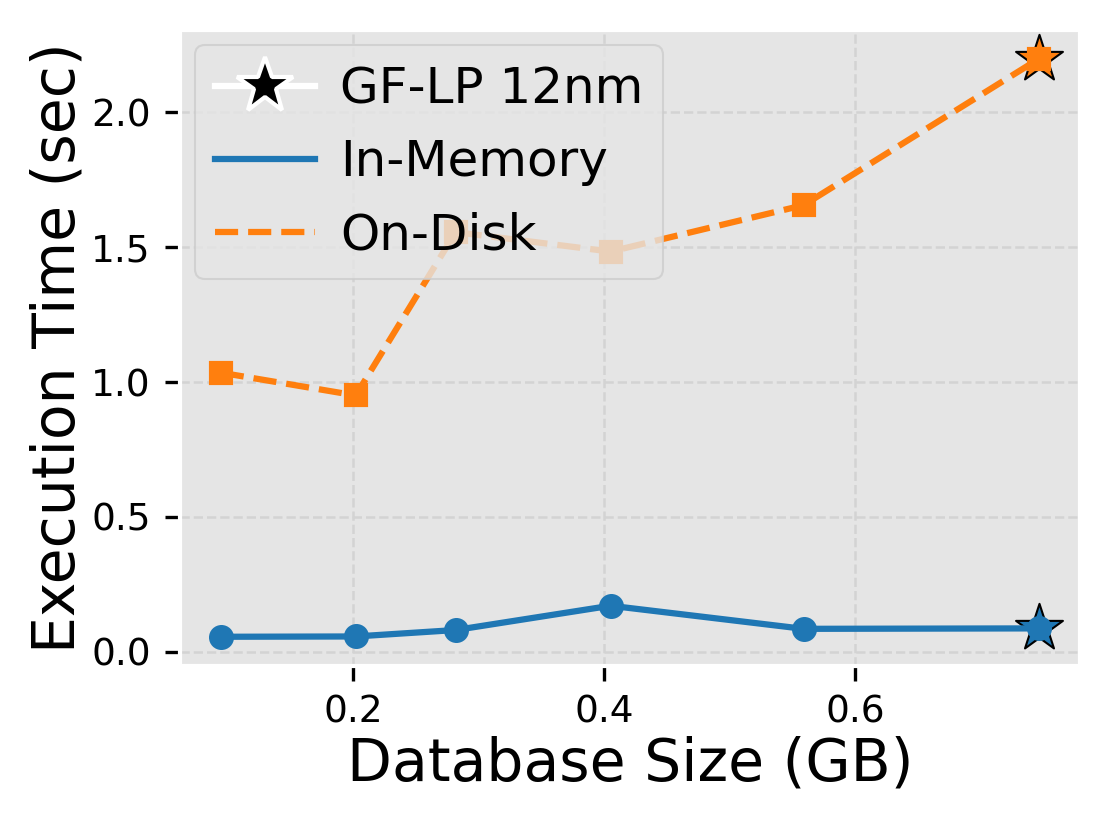}
        \vspace{-6mm}
        \caption{GF12nm LP SCLs}
        \label{fig:gf12}
    \end{subfigure}
\caption{PDK database query scaling: (a) Dynamic memory loading of PDK partitions based on dispatcher agent, (b) Query execution time vs. database size for Skywater 130nm PDK for in-memory and on-disk storage, (c) Scalability analysis for GF 12nm low power libraries.}

    \label{fig:db_scaling}

\vspace{-2mm}
\end{figure}

\textbf{\emph{Dispatcher:}} The \emph{dispatcher} routes user questions to the appropriate sources in the database. For PDK queries, it directs the question to the relevant standard cell library $\mathcal{L^\prime}$, library view $\mathcal{V^\prime}$, and operating conditions $\mathcal{C^\prime}$. For design queries, the \emph{dispatcher} routes the question to the relevant physical design stage $\mathcal{S^\prime}$. A key role of the dispatcher is resolving the structured redundancy in PDK and design data, ensuring that the \emph{Query-Generator} correctly filters tables by the relevant process corner, standard cell library, and physical design stage. Moreover, the \emph{dispatcher}'s output is used to ensure efficient scaling to larger PDKs through the dynamic loading of the relevant PDK partition into memory, in order to reduce the overhead of querying large databases as shown in Fig.~\ref{fig:db_scaling}. By leveraging in-memory storage rather than traditional disk-based retrieval, query latency is significantly reduced, as demonstrated by the execution time analysis for Skywater 130nm PDK in Fig.~\ref{fig:sky130} and GF 12nm low power libraries in Fig.~\ref{fig:gf12}. The function of the dispatcher agent is described in Eq.~\ref{eq:dispatcher}, where $\mathcal{Q}$ represents user question and $f_{dispatcher}(.|\mathcal{M})$ represents prompting the LLM agent $\mathcal{M}$ with specified inputs to make the routing decision. For PDK queries, the inputs are the available libraries $\mathcal{L}$ and library views $\mathcal{V}$, while for design queries, the inputs are the available physical design stages $\mathcal{S}$.

\vspace{-2.2mm}
\begin{equation}
\footnotesize
\begin{cases}
(\mathcal{L^\prime}, \mathcal{V^\prime}, \mathcal{C^\prime}) = f_{\text{\emph{dispatcher}}}(\mathcal{Q}, \mathcal{L}, \mathcal{V} \mid \mathcal{M}), & \text{if } \mathcal{DB} = \mathcal{PDK} \\
\mathcal{S^\prime} = f_{\text{\emph{dispatcher}}}(\mathcal{Q}, \mathcal{S} \mid \mathcal{M}) & \text{if } \mathcal{DB} =  Design
\end{cases}
\label{eq:dispatcher}
\end{equation}

\textbf{\emph{Selector:}} The \emph{selector} agent narrows the database schema to include only relevant tables (for the PDK relational database) or nodes (for the design graph database). For PDK queries, it filters the tables based on the routed view $\mathcal{V^\prime}$, retaining only those relevant to this view. The filtered schema $\mathcal{T^\prime}$ is then passed to the \emph{selector} agent for further refinement, selecting only tables directly related to the user question. For design queries, the selector identifies the most relevant nodes. This schema reduction simplifies text-to-SQL and text-to-Cypher tasks, allowing the \emph{Query-Generator} to focus on pertinent schema elements, improving both efficiency and accuracy in query generation. The function of the selector agent is described in Eq.~\ref{eq:selector}, where $f_{selector}(.|\mathcal{M})$ represents prompting LLM $\mathcal{M}$ with the specified inputs to make schema selection decisions.

\begin{equation}
\vspace{-0mm}
\footnotesize
\begin{cases}
\mathcal{T^{\prime\prime}} = f_{\text{\emph{selector}}}(\mathcal{Q}, \mathcal{T^\prime} \mid \mathcal{M}),  \mathcal{T^\prime} = \{ t \in \mathcal{T} \mid \mathcal{V^\prime} \} & \text{if } \mathcal{DB} = \mathcal{PDK} \\
\mathcal{N^{\prime}} = f_{\text{\emph{selector}}}(\mathcal{Q}, \mathcal{N} \mid \mathcal{M}), & \text{if } \mathcal{DB} = Design
\end{cases}
\label{eq:selector}
\end{equation}

\textbf{\emph{Query-Generator:}} The \emph{Query-Generator} constructs database queries to retrieve information needed to answer the user question. It relies on the dispatcher agent’s output to resolve structured redundancy in PDK and design data, ensuring that queries are constructed with the correct filtering conditions for the relevant entries. For PDK queries, it builds a SQL query based on the selected schema $\mathcal{T^{\prime\prime}}$, routed standard cell library $\mathcal{L^\prime}$, and operating conditions $\mathcal{C^\prime}$. For design queries, it constructs a Cypher query based on the selected nodes $\mathcal{N^\prime}$ and routed design stage $\mathcal{S^\prime}$. The \emph{Query-Generator} uses a chain of thought decomposition approach, where the LLM breaks down the user question into sub-questions, generating corresponding sub-queries for each~\cite{pourreza2024din}. These sub-queries are then combined into a final query, enhancing accuracy for complex questions. The \emph{Query-Generator} function is described in Eq.~\ref{eq:generator}, where $f_{generator}(.|\mathcal{M})$ represents prompting the LLM agent $\mathcal{M}$ with the specified inputs to generate the final database query.

\vspace{-3.2mm}
\begin{equation}
\fontsize{7.9}{10.9}\selectfont
Query = 
    \begin{cases}
        \mathcal{SQL} =  f_{\text{\emph{generator}}}(\mathcal{Q}, \mathcal{T^{\prime\prime}}, \mathcal{L^\prime}, \mathcal{C^\prime} \mid \mathcal{M}), & \text{if } \mathcal{DB} = \mathcal{PDK} \\
        Cypher = f_{\text{\emph{generator}}}(\mathcal{Q}, \mathcal{N^\prime}, \mathcal{S^\prime} \mid \mathcal{M}), & \text{if } \mathcal{DB} = Design
    \end{cases}
\label{eq:generator}
\end{equation}

\textbf{\emph{Refiner:}} The \emph{Refiner} fixes any operational errors that arise from the execution of generated queries, such as syntax issues, through an iterative process with a maximum of $n$ iterations. To ensure secure handling of proprietary PDK and design data, the raw database results are encrypted before being passed between API-based agents. Once the refinement is complete, the encrypted database $\mathcal{R}$ is passed to the \emph{Planner} agent, to indicate the end of the action execution. The function of the refiner is described in Eq.~\ref{eq:refiner}.

\begin{equation}
\fontsize{7.9}{10.9}\selectfont
\begin{aligned}
    \mathcal{R}, Error &= Execute(Query, \mathcal{DB})
    \\
    Query^\prime &= f_{\text{\emph{refiner}}}^{(n)}(\mathcal{Q}, Query, Error \mid \mathcal{M}) 
\end{aligned}
\label{eq:refiner}
\end{equation}

\textbf{\emph{Interpreter:}} The \emph{Interpreter} agent is triggered by the \emph{Planner} when the $interpret$ action is invoked. It processes the raw database $\mathcal{R}$ results into the final answer $\mathcal{A}$. Prior agents only have access to the database schemas and thus can be powered by API-based LLMs, however \emph{Interpreter} must access the raw database results in order to formulate the final answer. To ensure the secure handling of proprietary PDK and design data, the \emph{Interpreter} runs locally with a trusted LLM on the user’s side.




\section{Experimental Results}
\label{exp-results}

We implemented the framework using LangGraph~\cite{langgraph}. Our experiments were conducted using OpenAI's GPT-4 Turbo model (\textit{gpt-4-turbo-2024-04-09}) and the DeepSeek-V3 model (\textit{deepseek-v3}) via their respective APIs, as well as Meta's \textit{Llama-3.3-70B} model, which we ran locally. We set the temperature to 0 across all experiments to ensure deterministic outputs.

We used the open-source Skywater 130nm Process Design Kit (PDK)\cite{skywater}, which includes six Standard Cell Libraries (SCLs). We converted the PDK files into a SQL database based on our proposed schema. This yielded 19 tables with $39,576$ cell entries and 13,874,290 timing entries, totaling \SI{1.1}{\giga\byte} of data. For the design, we used a picorv CPU containing $50,637$ cells and $11,070$ nets. We parsed the design files using OpenRoad’s OpenDB\cite{Tutu2019openroad} and stored the information in a Neo4j graph database~\cite{neo4j_graph_database} per our schema. The resulting graph database comprises $605,563$ nodes and $2,582,105$ edges, creating a robust platform for evaluating our framework.

\subsection{Evaluation Setup}
\begin{table}[!t]
\centering
\caption{Evaluation set statistics, showing the number of queries and clause occurrences.}
\vspace{-2mm}
\resizebox{\linewidth}{!}{ 
\begin{threeparttable}
\begin{tabular}{@{}l@{\hspace{0.7em}}ccc@{\hspace{0.7em}}c@{\hspace{0.9em}}c@{\hspace{1em}}c@{}}
\toprule
\textbf{Statistic} & \multicolumn{3}{c}{\textbf{PDK}} & \multicolumn{1}{c}{\textbf{Design}\tnote{1}} & \textbf{Cross-Database\tnote{*}} & \textbf{Total} \\
\cmidrule(lr){2-4}
& \textbf{TechLEF} & \textbf{LEF} & \textbf{Liberty} & \multicolumn{1}{c}{ \textbf{(DEF)}} & & \\
\midrule
Total Questions & 23 & 23 & 28 & 35 & 8 & 117 \\
\midrule
\multicolumn{7}{@{}l}{\textbf{Clause Occurrences in Ground Truth Queries:}} \\
JOIN & 3 & 10 & 7 & - & 5 & 25 \\
ORDER BY & 2 & 7 & 8 & 11 & 6 & 34 \\
WHERE & 26 & 40 & 79 & 11 & 10 & 166 \\
GROUP BY & 6 & 5 & 4 & - & 4 & 19 \\
Aggregation Functions & 17 & 14 & 26 & 18 & 8 & 83 \\
Sub-queries & 3 & 17 & 41 & 4 & 7 & 72 \\
\midrule
Avg. Query Length (chars) & 161 & 189 & 327 & 129 & 256 & 215 \\
Max. Query Length (chars) & 452 & 661 & 761 & 258 & 893 & 975 \\
\bottomrule
\end{tabular}
\begin{tablenotes}
\footnotesize
\item[1] Design uses Cypher queries, while PDK uses SQL.
\item[*] Cross-Database questions require querying both the PDK and design database.
\end{tablenotes}
\end{threeparttable}
} 
\label{tab:query_stats}
\end{table}

We created an evaluation set of $117$ user questions and corresponding database queries, spanning a range of complexities from simple single-table selections to complex multi-table joins with sub-queries. Table~\ref{tab:query_stats} summarizes the occurrences of SQL and Cypher clauses in the evaluation set. We evaluated the performance of our framework using Execution Accuracy (EX) and Valid Efficiency Score (VES)\cite{wang2023mac,li2024can,zhong2020semantic}. Execution Accuracy (EX) measures the framework's ability to generate database queries that yield results matching the ground truth. EX is defined in Eq. \ref{eq:ex}, where $N$ is the total number of questions, $V_i$ is the result from the ground truth query, and $\hat{V_i}$ is the result from the LLM-generated query. The indicator function $1\!\!1(.)$ equals $1$ if the predicted result matches the ground truth and $0$ otherwise.

\vspace{-0mm}
\begin{equation}
\footnotesize
\label{eq:ex}
EX = \frac{\sum_{i=1}^N 1\!\!1(V_i, \hat{V}_i)}{N}, \quad 
1\!\!1(V_i, \hat{V}_i) = 
\begin{cases}
1, & \text{if } V_i = \hat{V}_i \\
0, & \text{if } V_i \ne \hat{V}_i
\end{cases}
\end{equation}

The Valid Efficiency Score (VES) evaluates the efficiency of correctly generated queries by comparing their execution time with the ground truth. VES, defined in Eq.~\ref{eq:ves}, uses $R(.)$ to represent relative efficiency and $E(.)$ for execution time. This score offers insights into both the accuracy and computational efficiency of the generated queries.

\vspace{-3mm}
\begin{equation}
\footnotesize
\label{eq:ves}
VES = \frac{\sum_{i=1}^N 1\!\!1(V_i, \hat{V}_i) \cdot R(Y_i, \hat{Y}_i)}{N}, \quad 
R(Y_i, \hat{Y}_i) = \sqrt{\frac{E(Y_i)}{E(\hat{Y}_i)}}
\end{equation}
\vspace{-1mm}

\subsection{Case Studies}
\definecolor{verylightblue}{RGB}{220,235,250}

\begin{figure}[!t]
    \centering
    \begin{subfigure}{\linewidth}
        \centering
        \begin{tcolorbox}[
            colback=yellow!15,
            fonttitle=\bfseries,
            width=\linewidth,
            boxrule=0.1mm,
            arc=1mm,
            fontupper={\fontsize{7}{8}\selectfont},
            left=1pt,
            right=0pt,
            top=0pt,
            bottom=-1.4pt
        ]
            {
            \textbf{(a) User Question}\\
            \fontsize{7}{7}\selectfont
            Compare the leakage power, clock pin capacitance, area, and average propagation delay of the flip-flop cells with drive strength 4 across the different libraries in the PDK
            }
        \end{tcolorbox}
        \vspace{-4mm} 
       \begin{tcolorbox}[
        colback=green!10,
        title=,
        fonttitle=\bfseries,
        width=\linewidth,
        boxrule=0.1mm,
        arc=1mm,
            fontupper={\fontsize{7}{8}\selectfont},
        left=0pt,
        right=0pt,
        top=0pt,
        bottom=-6pt
    ]
        \begingroup
        \setlength{\parindent}{0pt}
        \setlength{\parskip}{0pt}
        \textbf{Generated SQL}
        \vspace{-1.3mm}
        \begin{lstlisting}[style=SQLStyle, caption=]
SELECT OC.Cell_Library, C.Name, C.Cell_Leakage_Power, C.Area, 
IP.Clock_Pin_Capacitance, TV.Avg_Propagation_Delay FROM 
Operating_Conditions OC JOIN Cells C ON OC.Condition_ID = 
C.Condition_ID LEFT JOIN Input_Pins IP ON C.Cell_ID = IP.Cell_ID AND 
IP.Clock = True LEFT JOIN (SELECT Cell_ID, AVG(Average_Delay) AS 
Avg_Propagation_Delay FROM Timing_Values GROUP BY Cell_ID) TV ON
C.Cell_ID = TV.Cell_ID WHERE OC.Temperature = 25.0 AND 
OC.Voltage = 1.8 AND C.Is_Flip_Flop = True AND C.Drive_Strength = 4
\end{lstlisting}
        \endgroup
    \end{tcolorbox}
    \vspace{-4mm} 
    \begin{tcolorbox}[
        colback=cyan!10,
        title=,
        fonttitle=\bfseries,
        width=\linewidth,
        boxrule=0.1mm,
        arc=1mm,
        fontupper={\fontsize{7}{8}\selectfont},
        left=1pt,
        right=0pt,
        top=0pt,
        bottom=-1.4pt
    ]
        \begingroup
        \setlength{\parindent}{0pt}
        \setlength{\parskip}{0pt}
        \textbf{\emph{ChipXplore} Answer (49 Seconds)}

Here's a breakdown for each parameter across libraries:

1. Leakage Power (in nW)

- sky130\_fd\_sc\_hd: Ranges from 0.008783073 to 0.01396726. ....

2. Clock Pin Capacitance (in pF)

- sky130\_fd\_sc\_hd: Ranges from 0.001774 to 0.002439. .....

4. Average Propagation Delay (in ns)

- sky130\_fd\_sc\_hd: Ranges from 0.552643 to 0.754195.

- sky130\_fd\_sc\_hs: Ranges from 0.273567 to 0.303940 ...

\#\#\# Summary

-Average Propagation Delay: The `sky130\_fd\_sc\_hs` 
shows the lowest delays, indicating potentially faster 
performance, whereas `sky130\_fd\_sc\_hd` shows higher delays, 
which might be a trade-off for lower power consumption 
or smaller area. .....

        \endgroup
    \end{tcolorbox}
    \vspace{-2.4mm}
    \vspace{0.5mm}
    \end{subfigure}
    \begin{subfigure}{\linewidth}
        \centering
        \begin{tcolorbox}[
            colback=yellow!15,
            fonttitle=\bfseries,
            width=\linewidth,
            boxrule=0.1mm,
            arc=1mm,
            fontupper={\fontsize{7}{8}\selectfont},
            left=1pt,
            right=0pt,
            top=0pt,
            bottom=-1.8pt
        ]
            {
            \textbf{(b) User Question}\\
            \fontsize{7}{7}\selectfont
List nets exceeding thresholds: routed length > 190 microns or fanout > 20, excluding power and ground nets. Analyze the nets' name, fanout, routed length, and metal layers.
}
        \end{tcolorbox}
        \vspace{-4mm} 
       \begin{tcolorbox}[
        colback=green!10,
        title=,
        fonttitle=\bfseries,
        width=\linewidth,
        boxrule=0.1mm,
        arc=1mm,
        fontupper={\fontsize{7}{8}\selectfont},
        left=1pt,
        right=0pt,
        top=0pt,
        bottom=-6pt
    ]
        \begingroup
        \setlength{\parindent}{0pt}
        \setlength{\parskip}{0pt}
        \textbf{Generated Cypher}
        \vspace{-1.3mm}
        \begin{lstlisting}[style=SQLStyle, caption=]
MATCH (d:Design {stage: 'routing'})-[:CONTAINS_NET]->(n:Net) WHERE 
n.signal_type <> 'POWER' AND n.signal_type <> 'GROUND' AND 
(n.routed_length > 190 OR n.fanout > 20) RETURN n.net_name, n.fanout, 
n.routed_length, [s IN [(n)-[:ROUTED_ON]->(seg:Segment) | seg.layer] | s] 
\end{lstlisting}
        \endgroup
    \end{tcolorbox}
    \vspace{-4mm} 
    \begin{tcolorbox}[
        colback=cyan!10,
        title=,
        fonttitle=\bfseries,
        width=\linewidth,
        boxrule=0.1mm,
        arc=1mm,
        fontupper={\fontsize{7}{8}\selectfont},
        left=1pt,
        right=0pt,
        top=0pt,
        bottom=0pt
    ]
        \begingroup
        \setlength{\parindent}{0pt}
        \setlength{\parskip}{0pt}
        \textbf{\emph{ChipXplore} Answer (24 Seconds)}

Here is a breakdown of the analyzed data for each net:

1. Net Name: clknet\_2\_3\_\_leaf\_clk\_i

   - Fanout: 8.0
 
   - Routed Length: 246.54 microns
   
   - Metal Layers: ['met3', 'met1', 'met2']
....

4. Net Name: \_1029\_

   - Fanout: 6.0

   - Routed Length: 192.75 microns
   
   - Metal Layers: ['met3', 'met2', 'met1']
 
 \endgroup
    \end{tcolorbox}
    \vspace{-2.2mm}
    \end{subfigure}

     \begin{subfigure}{\linewidth}
        \centering
        \begin{tcolorbox}[
            colback=yellow!15,
            fonttitle=\bfseries,
            width=\linewidth,
            boxrule=0.1mm,
            arc=1mm,
            fontupper={\fontsize{7}{8}\selectfont},
            left=1pt,
            right=0pt,
            top=0pt,
            bottom=-2pt
        ]
            {
            \textbf{(c) User Question}\\
            \fontsize{7}{7}\selectfont
    What would be the total increase in area and static power if we replaced buffer cells connected to the 'rst\_i' net in the current design with the buf\_2 cell from the high speed library in the PDK ?            }
        \end{tcolorbox}
        \vspace{-4mm} 
       \begin{tcolorbox}[
        colback=green!10,
        title=,
        fonttitle=\bfseries,
        width=\linewidth,
        boxrule=0.1mm,
        arc=1mm,
        fontupper={\fontsize{7}{8}\selectfont},
        left=0pt,
        right=0pt,
        top=0pt,
        bottom=-6pt
    ]
        \begingroup
        \setlength{\parindent}{0pt}
        \setlength{\parskip}{0pt}
        \textbf{Generated SQL}
        \vspace{-1.2mm}
        \begin{lstlisting}[style=SQLStyle, caption=]
SELECT Area, Cell_Leakage_Power AS Static_Power FROM Cells WHERE Name 
= 'sky130_fd_sc_hs__buf_2' AND Condition_ID = ( SELECT Condition_ID 
FROM Operating_Conditions WHERE Temperature = 25.0 AND Voltage = 1.8
AND Cell_Library = 'sky130_fd_sc_hs' );
        \end{lstlisting}
        \vspace{-1.2mm}
        \textbf{Generated Cypher}
        \vspace{-1.2mm}
        \begin{lstlisting}[style=SQLStyle, caption=]
MATCH (d:Design {stage: 'routing'})-[:CONTAINS_NET]->(n:Net {net_name 
:'rst_i'})-[:CONNECTED_TO]->(c:Cell) WHERE c.is_buf = true RETURN c
        \end{lstlisting}
        \endgroup
    \end{tcolorbox}
    \vspace{-4mm} 
    \begin{tcolorbox}[
        colback=cyan!10,
        title=,
        fonttitle=\bfseries,
        width=\linewidth,
        boxrule=0.1mm,
        arc=1mm,
        fontupper={\fontsize{7}{8}\selectfont},
        left=1pt,
        right=0pt,
        top=0pt,
        bottom=-2pt
    ]
        \begingroup
        \setlength{\parindent}{0pt}
        \setlength{\parskip}{0pt}
        \textbf{\emph{ChipXplore} Answer (52 Seconds)}

The existing buffer cell has an area of 3.7536 and a static 
power of 7.016e-5. The 'sky130\_fd\_sc\_hs\_\_buf\_2' cell has 
an area of 7.992 and a static power of 0.07495. The 
increase in area per cell is 7.992 - 3.7536 = 4.2384, and 
the increase in static power per cell is 
0.07495 - 7.016e-5. The total increase in area 
would be 4.2384 and the total increase in static power 
would be approximately 0.07495.
        \endgroup
    \end{tcolorbox}
    \end{subfigure}
    \vspace{-5mm}
\caption{Example queries demonstrating \emph{ChipXplore's} ability to retrieve and analyze design and PDK data: (a) comparing flip-flop characteristics across PDK libraries, (b) identifying nets exceeding fanout and length thresholds, and (c) evaluating power and area trade-offs for buffer replacements}
\label{fig:cases}
\vspace{-2mm}
\end{figure}

\definecolor{lightgray}{gray}{0.9}

\begin{table*}[!t]
    \centering
    \begin{threeparttable}

    \fontsize{7}{7}\selectfont
    \caption{\emph{ChipXplore} customized workflow versus generic in-context learning workflows. The table summarizes the Execution Accuracy (EX) and Valid Efficiency Score (VES) using different backbone models. A checkmark ($\checkmark$) indicates an open-source model, while a cross ($\times$) indicates a proprietary model. \textbf{Bolded} values highlight the best-performing backbone model for the given workflow. \colorbox{lightgray}{\textbf{Bolded}} values highlight the best-performing workflow + backbone model. \vspace{-1.6mm}}
    \begin{tabular}{
        p{1.5cm} | p{1.8cm} | c | 
        r r |  
        r r |  
        r r |  
        r r  
    }
        \toprule
        \centering \textbf{Workflow} & \textbf{Model} & \textbf{Open-Source} & \multicolumn{2}{c|}{\textbf{PDK (SQL)}} & \multicolumn{2}{c|}{\textbf{Design (Cypher)}} & \multicolumn{2}{c}{\textbf{Cross-Database}} & \multicolumn{2}{|c}{\textbf{Overall}} \\
        \cmidrule(lr){4-5} \cmidrule(lr){6-7} \cmidrule(lr){8-9} \cmidrule(lr){10-11}
        & \multicolumn{1}{c|}{\textbf{}} & \multicolumn{1}{c|}{} & {\textbf{EX} (\%)} & {\textbf{VES} (\%)} & {\textbf{EX} (\%)} &  {\textbf{VES} (\%)}  & {\textbf{EX} (\%)} &  {\textbf{VES} (\%)}  & {\textbf{EX} (\%)} &  {\textbf{VES} (\%)}  \\
        \midrule
        \multirow{4}{*}[0pt]{\centering \makecell{Vanilla-RAG}} 
        & \multicolumn{1}{l|}{\textsc{GPT-4}}    &  $\times$ & 44.62 & 40.83 & 37.00 & 30.22 & 0.0 & 0.0 & 38.81 & 34.73 \\ 
        & \multicolumn{1}{l|}{\textsc{Llama-3.3-70B}} & $\checkmark$ & 24.51 & 22.62 & 54.00 & 49.85  & 0.0 & 0.0 & 28.56 & 26.36 \\ 
        & \multicolumn{1}{l|}{\textsc{DeepSeek-V3}}  & $\checkmark$ & \textbf{46.82} & \textbf{44.29} & \textbf{71.00} & \textbf{68.33}  & 0.0 & 0.0 & \textbf{47.64} & \textbf{45.32} \\ 
        \midrule 
        \multirow{4}{*}[0pt]{\centering \makecell{ DIN-SQL~\cite{pourreza2024din}}} 
        & \multicolumn{1}{l|}{\textsc{GPT-4}}   & $\times$ & 29.34	& 23.79 &  30.00 & 32.22 & 0.0 & 0.0 & 	27.53 &	24.68 \\ 
        & \multicolumn{1}{l|}{\textsc{Llama-3.3-70B}} & $\checkmark$ & 26.12 &	18.59&  31.00	& 31.50 & 0.0 & 0.0 & 25.79 &	21.18 \\ 
        & \multicolumn{1}{l|}{\textsc{DeepSeek-V3}}  & $\checkmark$ & \textbf{ 63.41} & \textbf{63.44} & \textbf{55.00} & \textbf{53.87} & 0.0 & 0.0 & \textbf{56.56} & \textbf{56.24} \\ 
        \midrule 

        \multirow{4}{*}[0pt]{\centering \makecell{ MAC-SQL~\cite{wang2023mac}}} 
        & \multicolumn{1}{l|}{\textsc{GPT-4}}    &  $\times$ & 63.39 & 62.99 & 60.00 & 58.97 & 0.0 & 0.0 & 58.04	& 57.48 \\
        & \multicolumn{1}{l|}{\textsc{Llama-3.3-70B}} & $\checkmark$ & 60.86 & 66.92 & 66.00 & 68.91 & 0.0 & 0.0 & 58.24 & 62.94 \\ 
        & \multicolumn{1}{l|}{\textsc{DeepSeek-V3}}  & $\checkmark$ & \textbf{74.26} & \textbf{88.39} & \textbf{69.00} & \textbf{71.56} & 0.0 & 0.0 & \textbf{67.61} & \textbf{77.31} \\ 
        \midrule

        \multirow{4}{*}[0pt]{\centering \makecell{\textbf{ChipXplore} \\ \textbf{(Ours)}}}
        & \multicolumn{1}{l|}{\textsc{GPT-4}}    &  $\times$ & \cellcolor{lightgray}\textbf{97.30} & \cellcolor{lightgray}\textbf{104.47} & 94.28 & 93.14 & 75.00 & 73.00 &  94.87 & 99.07 \\  
        & \multicolumn{1}{l|}{\textsc{Llama-3.3-70B}} & $\checkmark$ & 71.65 & 77.45 & 97.00 &	94.84  & 87.50	& 85.20 & 80.31	& 81.54\\  
        & \multicolumn{1}{l|}{\textsc{DeepSeek-V3}}  & $\checkmark$ & \cellcolor{lightgray}\textbf{97.30} & 94.58 & \cellcolor{lightgray}\textbf{97.00} & \cellcolor{lightgray}\textbf{95.82} & \cellcolor{lightgray}\textbf{100.00} & \cellcolor{lightgray}\textbf{101.24} & \cellcolor{lightgray}\textbf{97.39}	& \cellcolor{lightgray}\textbf{95.32} \\

        & \multicolumn{1}{l|}{\textsc{\textbf{LLama-3.3-ChipXplore-70B\tnote{*}}}} & $\checkmark$ & 94.45 & 87.49  & 97.00 &	94.84 & 87.50	& 85.20 & 94.74 & 87.90   \\ 

        \bottomrule
    \end{tabular}
     \begin{tablenotes}
    \footnotesize
\item[*] The \texttt{LlaMa-3.3-ChipXplore-70b} model is finetuned on our SQL schema for PDK queries and generates SQL queries, while \texttt{LlaMa-3.3-70b} handles routing, schema selection, and cypher query generation.
\vspace{-4.8mm}
    \end{tablenotes}
    \label{tab:main_results}
    \end{threeparttable}

\end{table*}

We demonstrate our framework's utility through example tasks that query both the PDK and design databases. Fig.\ref{fig:cases} shows three cases: (a) querying the PDK database for performing cross-library comparison of flip-flop cell properties (e.g., area, speed, static power), critical for making informed trade-offs, (b) querying design database for analyzing nets that exceed certain thresholds of routed length and fanout, and (c) querying both databases to perform a quick evaluation of area and power trade-offs when substituting cells with high-speed variants. The average response time for these queries is $41.97$ seconds, highlighting the utility of \emph{ChipXplore} in accelerating time-intensive retrieval tasks that would require several minutes of manual effort, particularly for junior IC design engineers.

\subsection{Performance Evaluation}

To evaluate the performance of \emph{ChipXplore}, we compare its execution accuracy with other LLM-based workflows that are designed to query generic structured databases. Specifically, we compare to three in-context learning based baselines: (1) Vanilla-RAG which uses one monolithic LLM for directly translating user questions to database queries, (2) DIN-SQL ~\cite{pourreza2024din} which operates three stages that perform schema linking, query classification and decomposition, and self-correction, and (3) MAC-SQL~\cite{wang2023mac} which is a multi-agent framework that performs table selection, query decomposition, and refinement. 

The main difference between \emph{ChipXplore} and these frameworks is that \emph{ChipXplore} is capable of resolving the structured redundancy in PDK and design data through the \emph{dispatcher} agent and is capable of orchestrating reasoning over heterogeneous databases, thus can answer cross-database questions. As shown in Table.~\ref{tab:main_results},  \emph{ChipXplore} + \emph{DeepSeek-V3} achieves the highest overall execution accuracy, outperforming generic workflows by 29.78\% and highlighting the need for customization for handling hardware design data. 

We also explored instruction finetuning to improve smaller models' performance. Using \textit{GPT-4o-mini}, we generated a synthetic dataset of 2,224 text-to-SQL pairs (user questions and corresponding SQL queries) and finetuned the \emph{Llama-3.3-70b} model on our proposed SQL schema. The results in Table.~\ref{tab:main_results} show our finetuned \emph{Llama-3.3-ChipXplore-70b} model performed 22.8\% better than the base \emph{Llama-3.3-70b}. Furthermore, \emph{ChipXplore} with \emph{Llama-3.3-ChipXplore-70b} achieved comparable overall accuracy to both GPT-4 and DeepSeek-V3 models.

In Table.~\ref{tab:ablation}, we present an ablation study where we systematically removed individual agents from the \emph{ChipXplore + DeepSeek-V3} to assess each agent's contribution. We can see that the \emph{dispatcher} agent is essential as it helps subsequent agents correctly resolve the structured redundancy by filtering the data by the relevant library, operating conditions, and physical design stage.

\begin{figure}[!t]
    \centering
    \includegraphics[width=\linewidth]{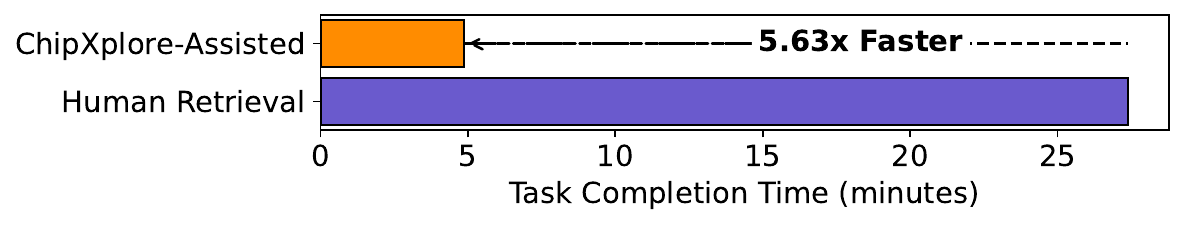}
    \vspace{-6mm}
\caption{Comparison of Task Completion Time: Manual Human Retrieval vs. \emph{ChipXplore}-Assisted Retrieval ($n=15$). }
\vspace{-2mm}
    \label{fig:study}
\end{figure}

To measure the productivity impact of \emph{ChipXplore}, we conducted a study with $15$ Electrical and Computer Engineering graduate students with prior knowledge of PDK formats, a demographic that represents the future IC engineer workforce. Participants completed $4$ retrieval tasks focused on PDK navigation twice; once manually and once using \emph{ChipXplore}. Fig.~\ref{fig:study} shows that the framework improved task completion time by 5.63× while reducing human errors by 5.25×, highlighting its utility in accelerating time-intensive tasks.

\begin{table}[t]
    \centering
\caption{Ablation study of \emph{ChipXplore} with different agents removed.}
\vspace{-2mm}
    \begin{tabular}{l|r|r}
        \toprule
        \textbf{Method} & \textbf{Execution Accuracy (EX \%)}  & \textbf{$\Delta (\%)$} \\
        \midrule
        \textbf{\emph{ChipXplore} (All Agents)} & \textbf{97.39\%} & - \\
        \midrule
        \emph{ChipXplore} wo-planner & 90.56\% & -6.83\% \\
        \emph{ChipXplore} wo-dispatcher & 68.46\% & -28.93\% \\
        \emph{ChipXplore} wo-selector & 83.52\% & -13.87\% \\
        \emph{ChipXplore} wo-decompose  & 89.94\% & -7.45\% \\
        \emph{ChipXplore} wo-refiner & 87.66\%  & -9.73\% \\
        \bottomrule
    \end{tabular}
    \label{tab:ablation}
    \vspace{-3mm}
\end{table}

\section{Conclusion}
\label{conclusion}

In this paper, we introduce \emph{ChipXplore}, a multi-agent, LLM-powered workflow that enables natural language interaction with Process Design Kits (PDKs) and hardware designs. PDK data is stored in a relational database, while hardware designs are organized in a graph database. \emph{ChipXplore} uses SQL and Cypher queries to retrieve relevant information and answer complex user questions, achieving 97.39\% execution accuracy across diverse queries, enhancing productivity by 5.63×, and reducing errors by 5.25×. \emph{ChipXplore} lays the foundation for building autonomous agents capable of tackling various physical design tasks that require PDK and hardware design awareness. 


\newpage

{\small
\bibliographystyle{ieeetr}
\bibliography{ref}
}

\newpage


\clearpage
\begin{appendix}

\subsection{Database Schema}
\label{database_schema}

We present comprehensive schema entity relationship diagrams that illustrate our approach to storing Physical Design Kit (PDK) files in relational databases. These include: Liberty files detailed in Fig.~\ref{liberty-er}, LEF files detailed in Fig.~\ref{lef-er} and TechnologyLEF files detailed in Fig.~\ref{tlef-er}. Additionally, Fig.~\ref{fig:def_db} displays the complete schema for storing DEF files in a graph database.

\begin{figure}[!h]
\centering
\resizebox{\columnwidth}{!}{%
  \tikzset{basic/.style={
      draw,
      rectangle split,
      rectangle split parts=2,
      rectangle split part fill={blue!20,white},
      text width=3.2cm,
      align=left,
      font=\footnotesize\itshape,
      inner xsep=2pt,
      inner ysep=2pt
  }}
  \begin{tikzpicture}[node distance=1cm and 0.3cm]
    \node[basic] (operating_conditions) {Operating\_Conditions
      \nodepart{second}
      \underline{Condition\_ID}\\
      Name\\
      Voltage\\
      Process\\
      Temperature\\
      Tree\_Type\\
      Cell\_Library};
    
    \node[basic, below=of operating_conditions] (cells) {Cells
      \nodepart{second}
      \underline{Cell\_ID}\\
      Name\\
      Drive\_Strength\\
      Area\\
      Cell\_Footprint\\
      Cell\_Leakage\_Power\\
      Driver\_Waveform\_Fall\\
      Driver\_Waveform\_Rise\\
      Is\_Buffer\\
      Is\_Inverter\\
      Is\_Flip\_Flop\\
      Is\_Scan\_Enabled\_Flip\_Flop\\
      Condition\_ID};
    
    \node[basic, right=of operating_conditions] (input_pins) {Input Pins
      \nodepart{second}
      \underline{Input\_Pin\_ID}\\
      Cell\_ID\\
      Input\_Pin\_Name\\
      Clock\\
      Capacitance\\
      Fall\_Capacitance\\
      Rise\_Capacitance\\
      Max\_Transition\\
      Related\_Power\_Pin\\
      Related\_Ground\_Pin};
    
    \node[basic, below=of cells] (output_pins) {Output Pins
      \nodepart{second}
      \underline{Output\_Pin\_ID}\\
      Cell\_ID\\
      Output\_Pin\_Name\\
      Function\\
      Max\_Capacitance\\
      Max\_Transition\\
      Power\_Down\_Function\\
      Related\_Power\_Pin\\
      Related\_Ground\_Pin};
    
    \node[basic, right=of input_pins] (input_pin_internal_powers) {Input Pin Internal Powers
      \nodepart{second}
      \underline{Internal\_Power\_ID}\\
      Input\_Pin\_ID\\
      Power\_Type\\
      Index\_1\\
      Value};
    
    \node[basic, right=of cells] (timing_values) {Timing Values
      \nodepart{second}
      \underline{Timing\_Value\_ID}\\
      Cell\_ID\\
      Output\_Pin\_ID\\
      Related\_Input\_Pin\\
      Input\_Transition\\
      Output\_Capacitance\\
      Fall\_Delay\\
      Rise\_Delay\\
      Average\_Delay\\
      Fall\_Transition\\
      Rise\_Transition};
    
    \draw[->, very thin] (cells) -- (operating_conditions);
    \draw[->, very thin] (input_pins) -- (cells);
    \draw[->, very thin] (output_pins) -- (cells);
    \draw[->, very thin] (input_pin_internal_powers) -- (input_pins);
    \draw[->, very thin] (timing_values) -- (cells);
    \draw[->, very thin] (timing_values) -- (output_pins);
  \end{tikzpicture}
}
\caption{Relational database schema for Liberty files.}
\label{liberty-er}
\end{figure}

\begin{figure}[!h]
\centering
\resizebox{\columnwidth}{!}{%
  \tikzset{basic/.style={
      draw,
      rectangle split,
      rectangle split parts=2,
      rectangle split part fill={blue!20,white},
      text width=2.7cm,
      align=left,
      font=\footnotesize\itshape,
      inner xsep=2pt,
      inner ysep=2pt
  }}
  \begin{tikzpicture}[node distance=1cm and 0.3cm]
    \node[basic] (macros) {Macros
      \nodepart{second}
      \underline{Macro\_ID}\\
      Name\\
      Class\\
      Foreign\_Name\\
      Origin\_X\\
      Origin\_Y\\
      Size\_Width\\
      Size\_Height\\
      Symmetry\\
      Site\\
      Cell\_Library};
    
    \node[basic, right=of macros] (pins) {Pins
      \nodepart{second}
      \underline{Pin\_ID}\\
      Macro\_ID\\
      Name\\
      Direction\\
      Use\\
      Antenna\_Gate\_Area\\
      Antenna\_Diff\_Area};
    
    \node[basic, right=of pins] (pinports) {Pin\_Ports
      \nodepart{second}
      \underline{Port\_ID}\\
      Pin\_ID\\
      Layer};
    
    \node[basic, below=of pinports] (pinportrects) {Pin\_Port\_Rectangles
      \nodepart{second}
      \underline{Rect\_ID}\\
      Port\_ID\\
      Rect\_X1\\
      Rect\_Y1\\
      Rect\_X2\\
      Rect\_Y2};
    
    \node[basic, below=of macros] (obstructions) {Obstructions
      \nodepart{second}
      \underline{Obstruction\_ID}\\
      Macro\_ID\\
      Layer};
    
    \node[basic, right=of obstructions] (obstructrects) {Obstruction\_Rectangles
      \nodepart{second}
      \underline{Obstruction\_Rect\_ID}\\
      Obstruction\_ID\\
      Rect\_X1\\
      Rect\_Y1\\
      Rect\_X2\\
      Rect\_Y2};
    
    \draw[->, very thin] (pins) -- (macros);
    \draw[->, very thin] (pinports) -- (pins);
    \draw[->, very thin] (pinportrects) -- (pinports);
    \draw[->, very thin] (obstructions) -- (macros);
    \draw[->, very thin] (obstructrects) -- (obstructions);
  \end{tikzpicture}
}
\caption{Relational database schema for LEF files.}
\label{lef-er}
\end{figure}

\begin{figure}[!t]
\centering
\resizebox{\columnwidth}{!}{%
  \tikzset{basic/.style={
      draw,
      rectangle split,
      rectangle split parts=2,
      rectangle split part fill={blue!20,white},
      text width=3.5cm,
      align=left,
      font=\footnotesize\itshape,
      inner xsep=2pt,
      inner ysep=2pt
  }}
  \begin{tikzpicture}[node distance=3cm and 0.3cm]
    \node[basic] (routing_layers) {Routing Layers
      \nodepart{second}
      \underline{Layer\_ID}\\
      Name \\  Type \\ Direction\\
      Pitch\_X \\ Pitch\_Y \\ Offset\_X \\ Offset\_Y\\
      Width \\
      Spacing \\ 
      Area\\
      Thickness \\ 
      Min\_Enclosed\_Area\\
      Edge\_Capacitance\\
      Capacitance\_Per\_SQ\_Dist\\
      Resistance\_Per\_SQ\\
      DC\_Current\_Density\_Avg\\
      AC\_Current\_Density\_Rms\\
      Maximum\_Density\\
      Density\_Check\_Window\_X \\
      Density\_Check\_Window\_Y\\
      Density\_Check\_Step\\
      Antenna\_Model\\
      Corner \\ Standard\_Cell\_Library};
    
    \node[basic, right=of routing_layers] (antenna_diff_side_area_ratios) {Antenna Diff Side\\Area Ratios
      \nodepart{second}
      \underline{Ratio\_ID}\\
      Routing\_Layer\_ID\\
      Type\\
      X1, \\ Y1\\ 
      X2, \\ Y2\\ 
      X3, \\ Y3\\ 
      X4, \\  Y4};
    
    \node[basic, right=of antenna_diff_side_area_ratios] (cut_layers) {Cut Layers
      \nodepart{second}
      \underline{Layer\_ID}\\
      Name, Type\\
      Width, Spacing\\
      Enclosure\_Below\_X \\
      Enclosure\_Below\_Y \\ 
      Enclosure\_Above\_X \\
      Enclosure\_Above\_Y\\
      Resistance\\
      DC\_Current\_Density\\
      Corner, \\ Standard\_Cell\_Library};
    
    \node[basic, below=of cut_layers] (antenna_diff_area_ratios) {Antenna Diff\\Area Ratios
      \nodepart{second}
      \underline{Ratio\_ID}\\
      Cut\_Layer\_ID\\
      Type\\
      X1 \\ Y1\\ 
      X2 \\ Y2\\ 
      X3 \\ Y3\\ 
      X4 \\ Y4};
    
    \node[basic, below=of routing_layers] (vias) {Vias
      \nodepart{second}
      \underline{Via\_ID}\\
      Name\\
      Upper\_Layer, \\ Lower\_Layer\\
      Corner\\
      Standard\_Cell\_Library};
    
    \node[basic, right=of vias] (via_layers) {Via Layers
      \nodepart{second}
      \underline{Via\_Layer\_ID}\\
      Via\_ID\\
      Layer\_Name\\
      Rect\_X1 \\ Rect\_Y1\\
      Rect\_X2 \\ Rect\_Y2};
    
    \draw[->, very thin] (antenna_diff_side_area_ratios) -- (routing_layers);
    \draw[->, very thin] (antenna_diff_area_ratios) -- (cut_layers);
    \draw[->, very thin] (via_layers) -- (vias);
  \end{tikzpicture}
}
\caption{Relational database schema for Technology LEF files.}
\label{tlef-er}
\end{figure}

\begin{figure}[!h]
\centering
\resizebox{\columnwidth}{!}{%
  \tikzset{basic/.style={
      draw,
      rectangle split,
      rectangle split parts=2,
      rectangle split part fill={yellow!20,white},
      text width=1.8cm,
      align=left,
      font=\footnotesize\itshape,
      inner xsep=2pt,
      inner ysep=2pt
  }}
  \begin{tikzpicture}[
      node distance=3cm and 2cm,
      every node/.style={font=\footnotesize},
      arrow/.style={->,>=latex,thick},
      labelstyle/.style={midway, font=\scriptsize, sloped}
  ]
  
    \node[basic] (design) {Design
      \nodepart{second}
      \underline{Design\_ID}\\
      Name\\
      Stage\\
      Die\_Area\\
      Xmin \\ 
      Ymin\\
      Xmax \\ 
      Ymax};
    
    \node[basic, below=of design] (port) {Port
      \nodepart{second}
      \underline{Port\_ID}\\
      Name\\
      Direction\\
      Signal\_Type\\
      Width\\ 
      Height\\
      Layer\\
      X \\
      Y};

       \node[basic, right=of design] (cell) {Cell
      \nodepart{second}
      \underline{Cell\_ID}\\
      Name\\
      Instance\_Name\\
      Orientation\\
      Dynamic\_Power\\
      X0 \\ 
      Y0 \\ 
      X1 \\ 
      Y1};

     \node[basic, right=of cell] (cellpin) {CellInternalPin
      \nodepart{second}
      \underline{Pin\_ID}\\
      Name\\
      Direction\\
      Transition\\
      Slack\\
      Rise\_arr \\ 
      Fall\_arr\\
      Capacitance};
      
    \node[basic, below =of cell] (net) {Net
      \nodepart{second}
      \underline{Net\_ID}\\
      Name\\
      Signal\_Type\\
      Capacitance\\
      Resistance\\
      Fanout\\
      Length};

    \node[basic, right=of net] (segment) {Segment
      \nodepart{second}
      \underline{Segment\_ID}\\
      Layer\\
      X0 \\ 
      Y0 \\ 
      X1 \\ 
      Y1};

    \draw[arrow] (design) -- node[labelstyle, above] {HAS\_PORT} (port);
    \draw[arrow] (design) -- node[labelstyle, above] {CONTAINS\_CELL} (cell);
    \draw[arrow] (design) -- node[labelstyle, above, bend left=20] {CONTAINS\_NET} (net);
    
    \draw[arrow] (cell) -- node[labelstyle, above] {HAS\_PIN} (cellpin);
    \draw[arrow, bend right=15] (cellpin) to node[labelstyle, above] {BELONGS\_TO} (cell);
    
    \draw[arrow, bend left=15] (cellpin) to node[labelstyle, below] {DRIVES\_NET} (net);
    \draw[arrow, bend right=-15] (net) to node[labelstyle, above] {CONNECTS\_TO} (cellpin);
    
    \draw[arrow, bend left=40] (port) to node[labelstyle, above] {DRIVES\_NET} (net);
    \draw[arrow, bend right=-40] (net) to node[labelstyle, below] {CONNECTS\_TO} (port);
    
    \draw[arrow] (net) -- node[labelstyle, above] {ROUTED\_ON} (segment);

  \end{tikzpicture}
}
\caption{Graph database schema for DEF files.}
\label{fig:def_db}
\end{figure}



\clearpage


\subsection{Error Analysis}
\label{eval_set}

\begin{figure}[!h]
    \centering
    \includegraphics[width=\columnwidth]{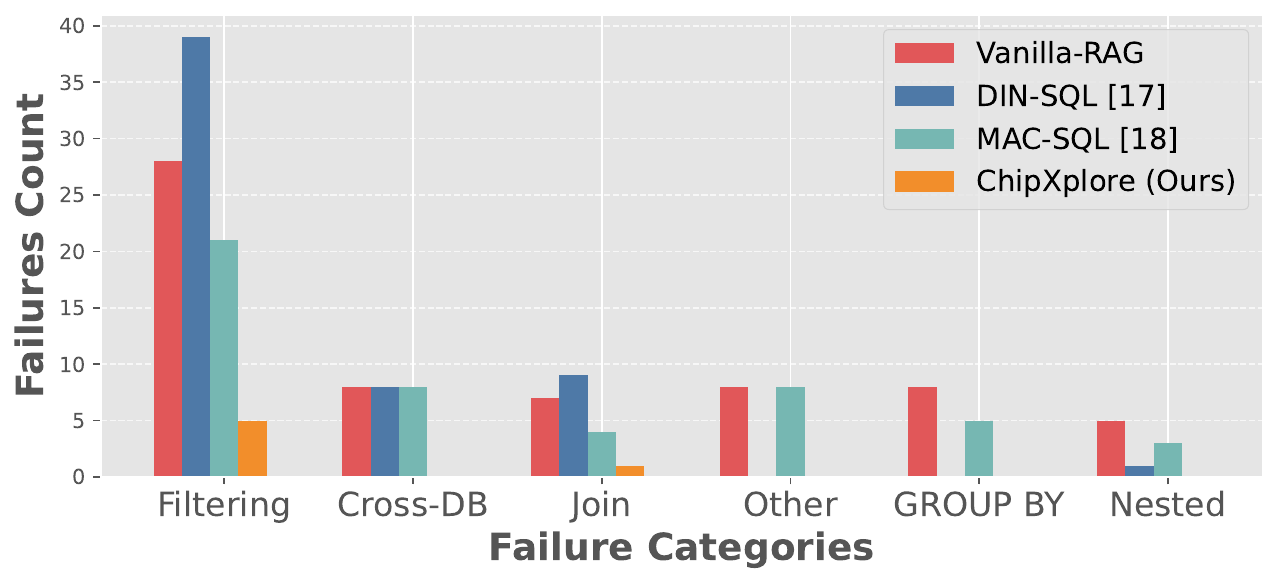}
    \vspace{-5mm}
    \caption{Breakdown of failure cases of \emph{ChipXplore} and three baseline workflows. }
    \label{fig:failure}
\end{figure}

Fig.~\ref{fig:failure} shows a breakdown of the failure categories of \emph{ChipXplore} versus three baseline workflows. We can see that the most common category is the filtering failure; this is mainly because baseline methods fail to correctly resolve the relevant standard cell library and operating conditions to which the question is referring to. \emph{ChipXplore} is capable of resolving structured redundancy through the \emph{dispatcher} agent, leading to significantly lower error rates in the filtering category. Table.~\ref{tab:failures_query_comparison} illustrates an example the filtering errors present in generic baseline workflows. 

\definecolor{darkgreen}{rgb}{0.0, 0.5, 0.0} 
\begin{table}[h]
    \centering
    \caption{Comparison of query generation across different baseline methods.}
    \begingroup
    \fontsize{9}{9}\selectfont
    \resizebox{\columnwidth}{!}{%
    \begin{tabular}{|p{10cm}|}
        \toprule
         \textbf{User Question 1:} How does the average leakage power vary across different cell libraries? \\
        \midrule
        \textbf{Ground Truth Query} 
        
        \texttt{SELECT oc.Cell\_Library, AVG(c.Cell\_Leakage\_Power) AS Average\_Leakage\_Power FROM Cells AS c JOIN Operating\_Conditions AS oc ON c.Condition\_ID = oc.Condition\_ID WHERE oc.Temperature = 25.0 AND oc.Voltage = 1.8 AND oc.Cell\_Library IN ('sky130\_fd\_sc\_hd', 'sky130\_fd\_sc\_hdll', 'sky130\_fd\_sc\_hs', 'sky130\_fd\_sc\_ls', 'sky130\_fd\_sc\_ms', 'sky130\_fd\_sc\_lp') GROUP BY oc.Cell\_Library} \\

        \midrule
        \textbf{Vanilla-RAG Query} 
        
        \texttt{SELECT AVG(Cell\_Leakage\_Power) FROM Cells WHERE Name LIKE 'sky130\_fd\_sc\_hd\_\_\%' AND Cell\_Leakage\_Power IS NOT NULL;} 

        \textcolor{red}{\textbf{Error} Wrong Cell Library Filters and Missing Operating Conditions Filters} \\

        \midrule
        \textbf{DIN-SQL~\cite{pourreza2024din}} 
        
        \texttt{SELECT oc.Cell\_Library, AVG(c.Leakage\_Power) FROM Cells c JOIN Operating\_Conditions oc ON c.Condition\_ID = oc.Condition\_ID GROUP BY oc.Cell\_Library}

        \textcolor{red}{\textbf{Error} Missing Operating Conditions Filters} \\

        \midrule
        \textbf{MAC-SQL~\cite{wang2023mac}} 
        
        \texttt{SELECT Cell\_Library, AVG(Cell\_Leakage\_Power) AS Avg\_Leakage\_Power FROM Cells JOIN Operating\_Conditions ON Cells.Condition\_ID = Operating\_Conditions.Condition\_ID GROUP BY Cell\_Library} 

        \textcolor{red}{\textbf{Error} Missing Operating Conditions Filters} \\
        
        \midrule
        \textbf{ChipXplore (Ours)}  
        
        \texttt{SELECT Cell\_Library, AVG(Cell\_Leakage\_Power) AS Average\_Leakage\_Power FROM Cells JOIN Operating\_Conditions ON Cells.Condition\_ID = Operating\_Conditions.Condition\_ID WHERE Operating\_Conditions.Temperature = 25.0 AND Operating\_Conditions.Voltage = 1.8 GROUP BY Cell\_Library} \\
        
        \textcolor{darkgreen}{\textbf{Correct}} \\

        \bottomrule
    \end{tabular}
}
    \endgroup
    \label{tab:failures_query_comparison}
\end{table}



\subsection{Synthetic text-to-SQL Dataset}
\label{synthetic_dataset}

We generated a synthetic dataset using \emph{gpt-4o-mini} for instruction fine-tuning on the text-to-SQL task. In total, we created 2,224 examples: 746 for the TechLef schema, 1061 for the lib schema, and 417 for the LEF schema. Fig.\ref{fig:lef_example} illustrates a sample from the LEF dataset. Fig.\ref{fig:lib_example} illustrates a sample from the liberty dataset. Fig.\ref{fig:techlef_example} illustrates a sample from the TechLef dataset.   

\begin{figure}[ht]
  \centering
    \begin{lstlisting}[language=json, frame=single, 
    xleftmargin=1em,
    xrightmargin=1em
    ]
"35": {
    "subtopic": {
        "id": 8,
        "name": "Cross-library Comparisons",
        "description": "Comparing cell dimensions across libraries and finding libraries with smallest/largest cells for given types"
    },
    "question": "What library has the largest cell height ?",
    "scl_library": [
        "HighDensity",
        "HighSpeed",
        "MediumSpeed",
        "LowPower",
        "LowSpeed",
        "HighDensityLowLeakage"
    ],
    "view": "Lef",
    "tables": [
        "Macros"
    ],
    "pvt_corner": null,
    "temp_corner": "",
    "voltage_corner": "",
    "techlef_op_cond": "",
    "query": "SELECT Cell_Library, MAX(Size_Height) AS Max_Height FROM Macros GROUP BY Cell_Library"
}
    \end{lstlisting}
  \caption{One example from the LEF synthetic text-to-sql dataset}
  \label{fig:lef_example}
\end{figure}

\begin{figure}[ht]
  \centering
    \begin{lstlisting}[language=json, frame=single, 
    xleftmargin=1em,
    xrightmargin=1em
    ]
"775": {
    "subtopic": {
    "id": 775,
    "name": "Average Leakage Power by Drive Strength",
    "description": "Analyzing how average leakage power varies with drive strength across libraries"
    },
    "question": "How does the average leakage power vary with drive strength across different libraries? Consider drive strengths 1, 2, 4, and 8.",
    "scl_library": [
      "HighDensity",
      "HighDensityLowLeakage",
      "HighSpeed",
      "MediumSpeed",
      "LowPower",
      "LowSpeed"
      ],
    "view": "Liberty",
    "tables": [
    "Operating_Conditions",
    "Cells"
    ],
    "pvt_corner": "tt_025C_1v80",
    "temp_corner": "25.0",
    "voltage_corner": "1.8",
    "techlef_op_cond": "",
    "query": "SELECT Operating_Conditions.Cell_Library, Cells.Drive_Strength, AVG(Cells.Leakage_Power) AS Avg_Leakage_Power FROM Cells JOIN Operating_Conditions ON Cells.Condition_ID = Operating_Conditions.Condition_ID WHERE Operating_Conditions.Temperature = 25.0 AND Operating_Conditions.Voltage = 1.8 AND Cells.Drive_Strength IN (1, 2, 4, 8) GROUP BY Operating_Conditions.Cell_Library, Cells.Drive_Strength ORDER BY Operating_Conditions.Cell_Library, Cells.Drive_Strength"
},
    \end{lstlisting}
  \caption{One example from the Liberty synthetic text-to-sql dataset}
  \label{fig:lib_example}
\end{figure}

\begin{figure}[ht]
  \centering
    \begin{lstlisting}[language=json, frame=single,    xleftmargin=1em,
    xrightmargin=1em]
"307": {
    "subtopic": {
        "id": 48,
        "name": "Layer Spacing Analysis",
        "description": "Examining spacing requirements across different layers"
    },
    "question": "Which routing layer has the smallest spacing to thickness ratio, and what is that ratio?",
    "scl_library": [
        "HighDensity"
    ],
    "view": "TechnologyLef",
    "tables": [
        "Routing_Layers"
    ],
    "pvt_corner": null,
    "temp_corner": "",
    "voltage_corner": "",
    "techlef_op_cond": [
        "nom"
    ],
    "query": "SELECT Name, Spacing, Thickness,  (Spacing/Thickness) AS Spacing_Thickness_Ratio FROM Routing_Layers WHERE Cell_Library = 'sky130_fd_sc_hd' AND Corner = 'nom' ORDER BY Spacing_Thickness_Ratio ASC LIMIT 1;"
},
    \end{lstlisting}
  \caption{One example from the TechLEF synthetic text-to-sql dataset}
  \label{fig:techlef_example}
\end{figure}

\subsection{User Study Statistics}
\label{eval_set}
\begin{figure}[!h]
    \centering
    \includegraphics[width=\columnwidth]{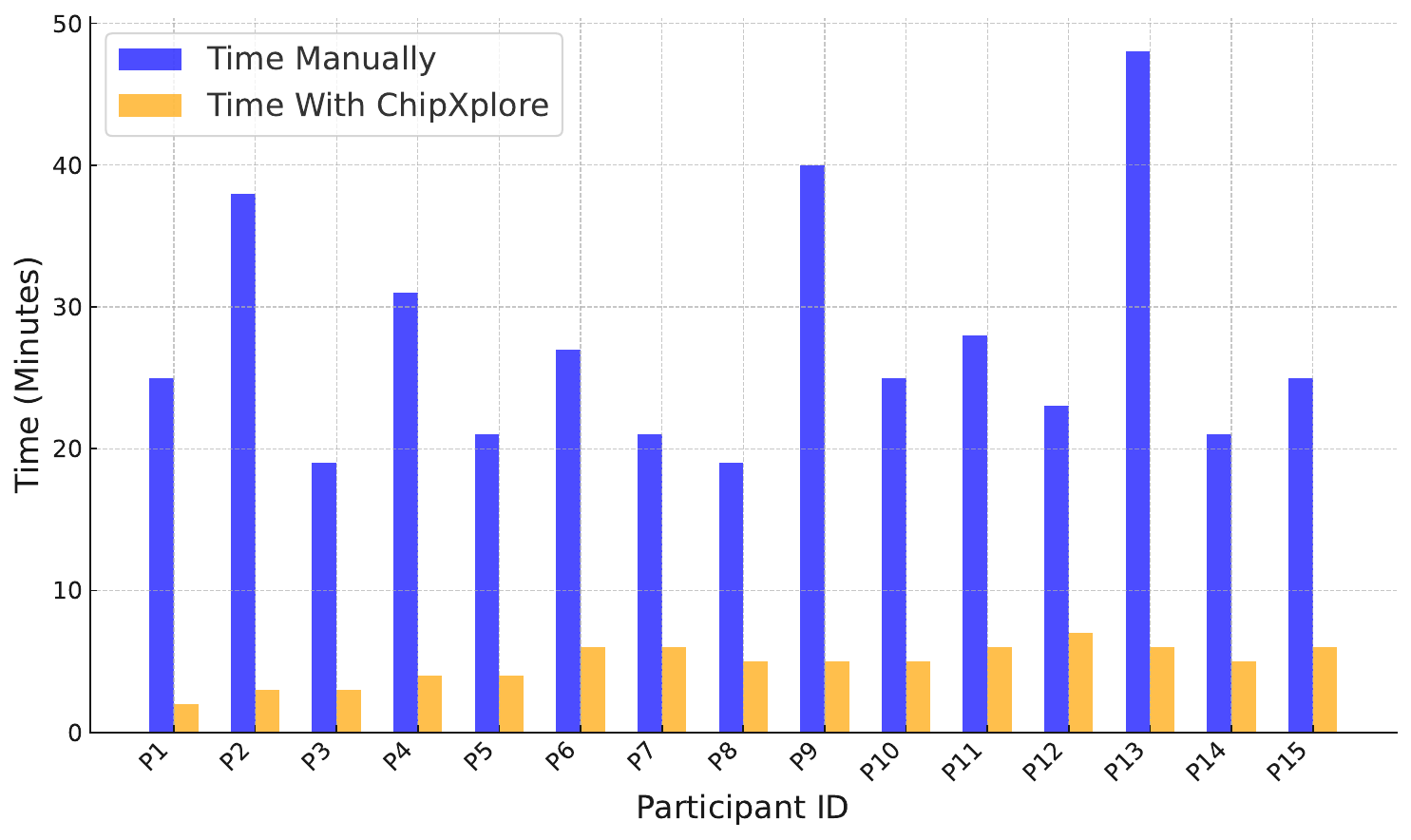}
    \caption{User study task completion time per participant.}
    \label{fig:user_study_by_part}
\end{figure}

Fig.~\ref{fig:user_study_by_part} shows the task completion time for the $15$ participants in the study. Table.~\ref{tab:bootstrap_stats} summarizes the statistics for the study such as the average task completion time and 95\% confidence intervals  using $10,000$ bootstrapped resamples. 

\begin{table}[h]
    \centering
    \caption{Mean task completion time with bootstrapped 95\% confidence intervals (10,000 resamples).}
    \renewcommand{\arraystretch}{1.2}
    \begin{tabular}{lrrr}
        \hline
        \textbf{Metric} & \textbf{Mean} & \textbf{95\% CI Lower} & \textbf{95\% CI Upper} \\
        \hline
        Time Manually & 27.40 & 24.12 & 32.18 \\
        Time With ChipXplore & 4.86 & 4.20 & 5.59 \\
        \hline
    \end{tabular}
    \label{tab:bootstrap_stats}
\end{table}

\newpage
\subsection{Cypher Interface versus EDA Tool API}
\label{eval_set}

We present additional experimental results comparing two approaches for retrieving design information: using the Cypher database interface versus relying on the EDA tool API. To conduct this comparison, we wrote equivalent OpenDB~\cite{Tutu2019openroad} TCL and Python scripts for the $35$ DEF questions in our evaluation set, allowing us to benchmark their execution times against Cypher queries. Table~\ref{tab:api_comparison} summarizes our findings. The comparison focuses on two key metrics: (1) token count, which measures query/script verbosity, and (2) execution run-time, which measures performance. Fig.~\ref{fig:token_comparison} provides an illustrative example where a Cypher query demonstrates significantly lower code complexity compared to the equivalent EDA tool API implementation.

\lstdefinestyle{SQLStyle2}{
    language=SQL,
    basicstyle=\fontsize{7}{7}\selectfont\ttfamily, 
    keywordstyle=\bfseries\color{blue},
    commentstyle=\itshape\color{gray},
    stringstyle=\color{red},
    morekeywords={MATCH, RETURN, WHERE, AND, OR}
}
\begin{figure}[!h]
    \centering
       \begin{tcolorbox}[
        colback=gray!10,
        title=,
        fonttitle=\bfseries,
        width=\linewidth,
        boxrule=0.1mm,
        arc=1mm,
        fontupper={\fontsize{8}{8}\selectfont},
        left=0pt,
        right=0pt,
        top=0pt,
        bottom=0pt
    ]
{       
    \textbf{User Question}\\
    \fontsize{8}{8}\selectfont
    What is the number of flip-flops cells in this rectangle (0, 0, 100, 100) ?
}

    \end{tcolorbox}

       \begin{tcolorbox}[
        colback=gray!10,
        title=,
        fonttitle=\bfseries,
        width=\linewidth,
        boxrule=0.1mm,
        arc=1mm,
        fontupper={\fontsize{8}{8}\selectfont},
        left=0pt,
        right=0pt,
        top=0pt,
        bottom=-6pt
    ]
        \textbf{Cypher}
        \vspace{-1.3mm}
        \begin{lstlisting}[style=SQLStyle2, caption=]
MATCH (c:Cell) WHERE c.x >= 0 AND c.x <= 100 
AND c.y >= 0 AND c.y <= 100 AND c.is\_seq = TRUE 
RETURN count(c) AS flop\_count;
\end{lstlisting}
    \end{tcolorbox}

   \begin{tcolorbox}[
        colback=gray!10,
        title=,
        fonttitle=\bfseries,
        width=\linewidth,
        boxrule=0.1mm,
        arc=1mm,
        fontupper={\fontsize{8}{8}\selectfont},
        left=0pt,
        right=0pt,
        top=0pt,
        bottom=-6pt
    ]
    \textbf{OpenDB Python Script}
    \vspace{-1.3mm}
    {\fontsize{7}{7}\selectfont 
    \begin{lstlisting}[caption=]
import odb
db = odb.dbDatabase.create()
chip = db.getChip()
top_block = chip.getBlock()
tech = db.getTech()
dbu_per_micron = tech.getDbUnitsPerMicron()
region_size_microns = 100
size_dbu = region_size_microns * dbu_per_micron
flop_count = 0
for inst in instances:
    bbox = inst.getBBox()
    llx, lly = bbox.xMin(), bbox.yMin()
    urx, ury = bbox.xMax(), bbox.yMax()
    if llx >= 0 and lly >= 0 and 
       urx <= size_dbu and ury <= size_dbu:
        in_region_count += 1
        master = inst.getMaster()
        if master.isSequential():
            flop_count += 1
print(f"Number of FFs: {flop_count}")

\end{lstlisting}
    }
\end{tcolorbox}

    \caption{Comparison of token count between Cypher query and equivalent OpenDB Python script for counting sequential cells in a specified region. The Cypher approach demonstrates a significantly less code complexity.}
    \label{fig:token_comparison}
\end{figure}

\begin{table}[htbp]
    \centering
    \caption{Performance comparison of Cypher query versus OpenDB APIs for retrieving design information.}
    \resizebox{\columnwidth}{!}{ 
    \begin{tabular}{l|r|r}
        \toprule
        \textbf{Interface} & \textbf{Avg. Execution Time (s)} & \textbf{Avg. Token Count} \\
        \midrule
        OpenDB's Python API & 0.534 & 338.057 \\
        \midrule
        OpenDB's TCL API & 0.601 & 394.709 \\
        \midrule
        Cypher & 0.019 & 40.828 \\
        \midrule
        \textbf{Improvements} & \textbf{96.44\%} & \textbf{87.92\%} \\
        \bottomrule
    \end{tabular}}
    \label{tab:api_comparison}
\end{table}

\end{appendix}

\end{document}